\documentclass[11pt,a4paper]{article}
\usepackage[hyperref]{naaclhlt2019}
\usepackage{times}
\usepackage{latexsym}
\usepackage{lipsum}
\usepackage{url}
\usepackage{multirow}
\usepackage{changepage}
\usepackage{amssymb}
\usepackage{bm}
\usepackage{amsmath}
\usepackage{graphicx}
\usepackage{booktabs}
\usepackage[hang,flushmargin]{footmisc}
\usepackage{subcaption}
\usepackage{cleveref}
\usepackage[nolist]{acronym}

\aclfinalcopy % Uncomment this line for the final submission
\def\aclpaperid{34} %  Enter the acl Paper ID here

\newcommand{\tbf}{\textbf}
\newcommand{\ul}{\underline}

\title{Gating Mechanisms for Combining Character
and Word-level Word Representations: An Empirical Study}

\author{Jorge A. Balazs \and Yutaka Matsuo \\
  Graduate School of Engineering\\
  The University of Tokyo\\
  \texttt{\{jorge, matsuo\}@weblab.t.u-tokyo.ac.jp} \\
}

\date{}

\begin{document}
\maketitle
\begin{abstract}

% In this paper we study how different ways of combining character and word-level
% representations affect the quality of sentence representations produced by a
% standard BiLSTM model with max pooling, and propose a fine-grained gating
% mechanism that does not rely on external features. We show that complementing
% word embeddings with character-level features is better on average than not
% doing so, that using gating mechanisms as opposed to simple concatenation
% induces better-quality sentence representations, and that our proposed mechanism
% performs the best in a significant number of evaluation tasks.

In this paper we study how different ways of combining character and word-level
representations affect the quality of both final word and sentence
representations. We provide strong empirical evidence that modeling characters
improves the learned representations at the word and sentence levels, and that
doing so is particularly useful when representing less frequent words. We
further show that a feature-wise sigmoid gating mechanism is a robust method for
creating representations that encode semantic similarity, as it performed
reasonably well in several word similarity datasets. Finally, our findings
suggest that properly capturing semantic similarity at the word level does not
consistently yield improved performance in downstream sentence-level tasks. Our
code is available at \url{https://github.com/jabalazs/gating}.

\end{abstract}

\section{Introduction}

% \usepackage{acronym}
% Needs to be declared in the caller to this file

\begin{acronym} % Give the longest label here so that the list is nicely aligned
\acro{BiGRU}{Bidirectional GRU}
\acro{BiLSTM}{Bidirectional LSTM}
\acro{BPTT}{Back Propagation Through Time}
\acro{CBOW}{Continuous Bag of Words}
\acro{CNN}{Convolutional Neural Network}
\acro{ELMo}{Embeddings from Language Models}
\acro{GRU}{Gated Recurrent Unit}
\acro{IEST}{Implicit Emotions Shared Task}
\acro{LSTM}{Long Short-Term Memory Network}
\acroindefinite{LSTM}{an}{a}
\acro{NLI}{Natural Language Inference}
\acro{NLP}{Natural Language Processing}
\acro{NN}{Neural Network}
\acro{RNN}{Recurrent Neural Network}
\acro{SST}{Stanford Sentiment Treebank}
\acro{TDNN}{Time Dilated Neural Network}
\acro{WASSA}{Workshop on Computational Approaches to Subjectivity, Sentiment
             \& Social Media Analysis}

\end{acronym}

\acrodefplural{BiLSTM}[BiLSTMs]{Bidirectional LSTMs}
\acrodefplural{CNN}[CNNs]{Convolutional Neural Networks}
\acrodefplural{GRU}[GRUs]{Gated Recurrent Units}
\acrodefplural{LSTM}[LSTMs]{Long Short-Term Memory Networks}
\acrodefplural{NN}[NNs]{Neural Networks}
\acrodefplural{RNN}[RNNs]{Recurrent Neural Networks}
\acrodefplural{TDNN}[TDNNs]{Time Dilated Neural Networks}

\hyphenation{VerbNet}

Incorporating sub-word structures like substrings, morphemes and characters
to the creation of word representations significantly increases their quality as
reflected both by intrinsic metrics and performance in a wide range of
downstream tasks~\citep{bojanowski2017enriching, luong2016achieving,
wu2016google, ling2015finding}.

The reason for this improvement is related to sub-word structures containing
information that is usually ignored by standard word-level models. Indeed, when
representing words as vectors extracted from a lookup table, semantically
related words resulting from inflectional processes such as \textit{surf},
\textit{surfing}, and \textit{surfed}, are treated as being independent from one
another\footnote{Unless using pre-trained embeddings with a notion of subword
information such as \texttt{fastText}~\citep{bojanowski2017enriching}}.
Further, word-level embeddings do not account for derivational processes
resulting in syntactically-similar words with different meanings such as
\textit{break}, \textit{breakable}, and \textit{unbreakable}. This causes
derived words, which are usually less frequent, to have lower-quality (or no)
vector representations.

Previous works have successfully combined character-level and word-level word
representations, obtaining overall better results than using only word-level
representations. For example~\citet{luong2016achieving} achieved
state-of-the-art results in a machine translation task by representing unknown
words as a composition of their characters. \citet{botha2014compositional}
created word representations by adding the vector representations of the words'
surface forms and their morphemes ($\overrightarrow{\text{perfectly}} =
\overrightarrow{perfectly} + \overrightarrow{perfect} + \overrightarrow{ly}$),
obtaining significant improvements on intrinsic evaluation tasks, word
similarity and machine translation. \citet{lample2016neural} concatenated
character-level and word-level representations for creating word
representations, and then used them as input to their models for obtaining
state-of-the-art results in Named Entity Recognition on several languages.

What these works have in common is that the models they describe first learn how
to represent subword information, at character~\citep{luong2016achieving},
morpheme~\citep{botha2014compositional}, or
substring~\citep{bojanowski2017enriching} levels, and then combine these learned
representations at the word level. The incorporation of information at a
finer-grained hierarchy results in higher-quality modeling of rare words,
morphological processes, and semantics~\citep{avraham2017interplay}.

There is no consensus, however, on which combination method works better in
which case, or how the choice of a combination method affects downstream
performance, either measured intrinsically at the word level, or extrinsically
at the sentence level.

In this paper we aim to provide some intuitions about how the choice of
mechanism for combining character-level with word-level representations
influences the quality of the final word representations, and the subsequent
effect these have in the performance of downstream tasks. Our contributions are
as follows:

\begin{itemize}

 \item We show that a feature-wise sigmoidal gating mechanism is the best at
     combining representations at the character and word-level hierarchies, as
     measured by word similarity tasks.

 \item We provide evidence that this mechanism learns that to properly model
     increasingly infrequent words, it has to increasingly rely on
     character-level information.

 \item We finally show that despite the increased expressivity of word
     representations it offers, it has no clear effect in sentence
     representations, as measured by sentence evaluation tasks.

\end{itemize}

\section{Background}

% \subsection{Combining Character-level and Word-level Word Representations}

We are interested in studying different ways of combining word representations,
obtained from different hierarchies, into a single word representation.
Specifically, we want to study how combining word representations (1) taken
directly from a word embedding lookup table, and (2) obtained from a function
over the characters composing them, affects the quality of the final word
representations.

Let $\mathcal{W}$ be a set, or vocabulary, of words with $|\mathcal{W}|$
elements, and $\mathcal{C}$ a vocabulary of characters with $|\mathcal{C}|$
elements. Further, let $\bm{x} = w_1, \ldots, w_n;\;w_i \in \mathcal{W}$ be a
sequence of words, and $\bm{c}^i = c^i_1, \ldots, c^i_m;\; c^i_j \in
\mathcal{C}$ be the sequence of characters composing $w_i$. Each token $w_i$ can
be represented as a vector $\bm{v}^{(w)}_i \in \mathbb{R}^d$ extracted directly
from an embedding lookup table $\bm{E}^{(w)} \in \mathbb{R}^{|\mathcal{W}|\times
d}$, pre-trained or otherwise, and as a vector $\bm{v}^{(c)}_i \in \mathbb{R}^d$
built from the characters that compose it; in other words, $\bm{v}^{(c)}_i =
f(\bm{c}^i)$, where $f$ is a function that maps a sequence of characters to a
vector.

The methods for combining word and character-level representations we study, are
of the form $G(\bm{v}^{(w)}_i, \bm{v}^{(c)}_i) = \bm{v}_i$ where $\bm{v}_i$ is
the final word representation.

% \section{From Characters to Character-level Word Representations}%
% \label{char_derivation}

\subsection{Mapping Characters to Character-level Word Representations}

The function $f$ is composed of an \emph{embedding} layer, an optional
\emph{context} function, and an \emph{aggregation} function.

The \textbf{embedding layer} transforms each character $c^i_j$ into a
vector $\bm{r}^i_j$ of dimension $d_r$, by directly taking it from a trainable
embedding lookup table $\bm{E}^{(c)} \in \mathbb{R}^{|\mathcal{C}|\times d_r}$.
We define the \emph{matrix} representation of word $w_i$ as $\bm{C}^i =
[\bm{r}^i_1, \ldots, \bm{r}^i_m], \enspace \bm{C}^i \in \mathbb{R}^{m \times
d_r}$.

The \textbf{context function} takes $\bm{C}^i$ as input and
returns a context-enriched matrix representation $\bm{H}^i = [\bm{h}_1^i,
\ldots, \bm{h}_m^i], \enspace \bm{H}^i \in \mathbb{R}^{m \times d_h}$, in which
each $\bm{h}^i_j$ contains a measure of information about its context, and
interactions with its neighbors. In particular, we chose to do this by feeding
$\bm{C}^i$ to a \ac{BiLSTM}~\citep{graves2005framewise,
graves2013speech}\footnote{Other methods for encoding the characters'
context, such as CNNs~\citep{kim2016character}, could also be used.}.

% \begin{equation*}
%     \begin{aligned}
%         f:\; &\mathbb{R}^{m \times d_r}&\to&\;\mathbb{R}^{d_r}\\
%              &\bm{C}^i&\mapsto&\;f(\bm{C}^i) = \bm{v}_i^{(c)}
%     \end{aligned}
% \end{equation*}

% It is also possible to use a CNN for adding context to the character-level
% vectors, as in \citep{kim2016character}

% \footnote{We chose to use a BiLSTM instead of alternative methods, such as a
% character-level CNN, not to deviate too much from previous works.}.

Informally, we can think of \iac{LSTM}~\citep{hochreiter1997lstm} as a function
$\mathbb{R}^{m \times d_r} \to \mathbb{R}^{m \times d_h}$ that takes
a matrix $\bm{C}=[\bm{r}_1,\ldots,\bm{r}_m]$ as input and returns a
context-enriched matrix representation $\bm{H} = [\bm{h}_1, \ldots, \bm{h}_m]$,
where each $\bm{h}_j$ encodes information about the previous elements $\bm{h}_1,
\ldots, \bm{h}_{j-1}$\footnote{In terms of implementation, the LSTM is applied
    iteratively to each element of the input sequence regardless of dimension
$m$, which means it accepts inputs of variable length, but we will use this
notation for the sake of simplicity.}.

A \ac{BiLSTM} is simply composed of 2 \acp{LSTM}, one that reads the input from
left to right (forward), and another that does so from right to left (backward).
The output of the forward and backward \acp{LSTM} are $\overrightarrow{\bm{H}} =
[\overrightarrow{\bm{h}}_1, \ldots, \overrightarrow{\bm{h}}_m]$ and
$\overleftarrow{\bm{H}} = [\overleftarrow{\bm{h}}_1, \ldots,
\overleftarrow{\bm{h}}_m]$ respectively. In the backward case the \ac{LSTM}
reads $\bm{r}_m$ first and $\bm{r}_1$ last, therefore $\overleftarrow{\bm{h}}_j$
will encode the context from $\overleftarrow{\bm{h}}_{j+1}, \ldots,
\overleftarrow{\bm{h}}_m$.

% To obtain a \emph{vector} representation for $w_i$, we have to reduce the sequence
% dimension of $\bm{C}^i$ or, in other words, aggregate the $\bm{r}^i_j$ vectors
% to obtain $\bm{v}^{(c)}_i$.

\begin{figure*}[!htbp]
    \centering
    \includegraphics[width=\textwidth]{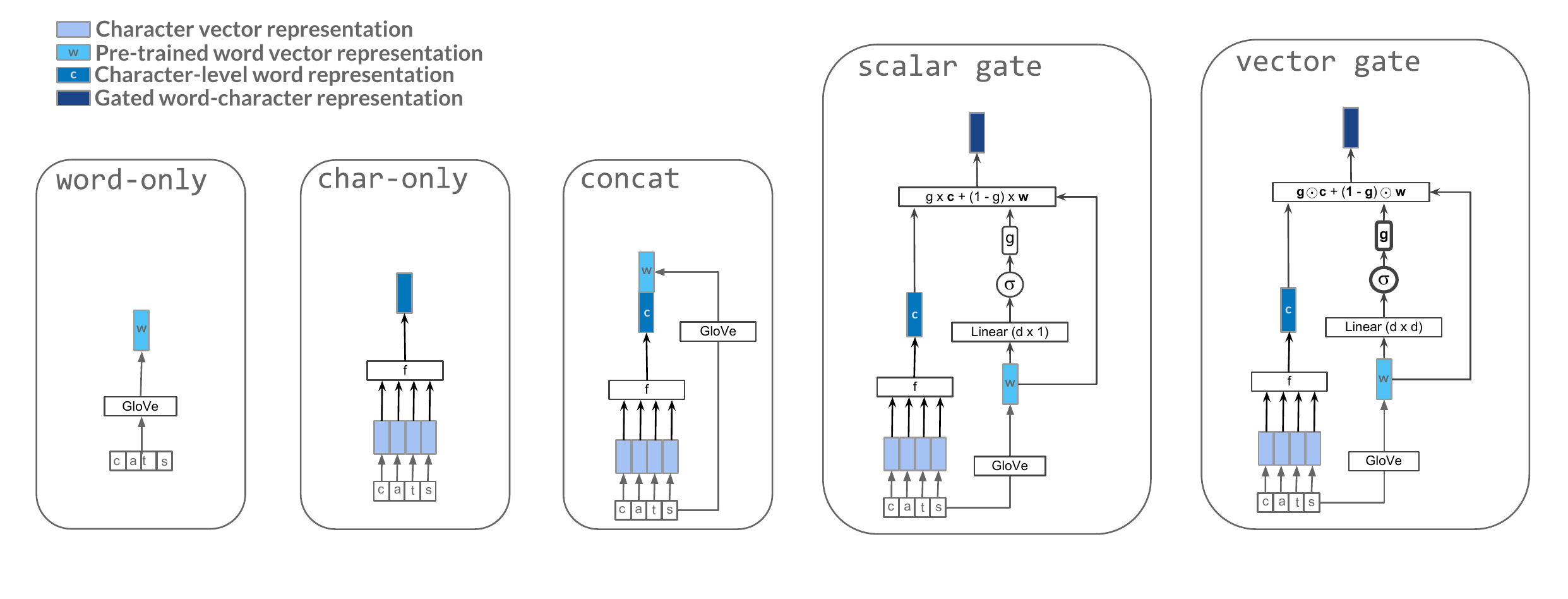}
    \caption{Character and Word-level combination methods.}
    \label{fig:methods}
\end{figure*}

The \textbf{aggregation function} takes the context-enriched matrix
representation of word $w_i$ for both directions, $\overrightarrow{\bm{H}^i}$
and $\overleftarrow{\bm{H}^i}$, and returns a single vector
$\bm{v}^{(c)}_i \in \mathbb{R}^{d_h}$. To do so we followed
\citet{miyamoto2016gated}, and defined the character-level representation
$\bm{v}^{(c)}_i$ of word $w_i$ as the linear combination of the forward and
backward last hidden states returned by the context function:
\begin{equation}
\label{eq:fw_bw_char}
    \bm{v}^{(c)}_i = \bm{W}^{(c)}[\overrightarrow{\bm{h}^i}_m;\overleftarrow{\bm{h}^i}_1] + \bm{b}^{(c)}
\end{equation}

\noindent
where $\bm{W}^{(c)}\in \mathbb{R}^{d_h \times 2d_h}$ and $\bm{b}^{(c)}
\in \mathbb{R}^{d_h}$ are trainable parameters, and $[\circ;\circ]$
represents the concatenation operation between two vectors.

% We chose this way of
% combining the last forward and backward hidden states over concatenation or
% addition because, unlike the former, it keeps a compatible dimension size with
% the word-level word representation $\bm{v}_i^{(w)} \in \mathbb{R}^{d_{lstm}}$,
% and its degree of expressivity is higher than the latter.

\subsection{Combining Character and Word-level Representations}\label{subsec:methods}

We tested three different methods for combining $\bm{v}^{(c)}_i$ with
$\bm{v}^{(w)}_i$: simple concatenation, a learned scalar
gate~\citep{miyamoto2016gated}, and a learned vector gate (also referred to as
feature-wise sigmoidal gate). Additionally, we compared these methods to two
baselines: using pre-trained word vectors only, and using character-only
features for representing words. See~\cref{fig:methods} for a visual
description of the proposed methods.

\noindent
\textbf{\texttt{word-only (w)}} considers only $\bm{v}^{(w)}_i$ and ignores
$\bm{v}^{(c)}_i$:
\begin{equation}
    \bm{v}_i = \bm{v}^{(w)}_i
\end{equation}

\noindent
\textbf{\texttt{char-only (c)}} considers only $\bm{v}^{(c)}_i$ and ignores
$\bm{v}^{(w)}_i$:
\begin{equation}
    \bm{v}_i = \bm{v}^{(c)}_i
\end{equation}

\noindent
\textbf{\texttt{concat (cat)}} concatenates both word and character-level
representations:
\begin{equation}
    \bm{v}_i = [\bm{v}^{(c)}_i;\bm{v}^{(w)}_i]
\end{equation}

\noindent
\textbf{\texttt{scalar gate (sg)}} implements the scalar gating mechanism described
by~\citet{miyamoto2016gated}:
\begin{align}
    g_i &= \sigma(\bm{w}^\top\bm{v}^{(w)}_i + b)\\
    \bm{v}_i &= g_i\bm{v}^{(c)}_i + (1 - g_i)\bm{v}^{(w)}_i\label{eq:scalar-gate}
\end{align}
\noindent
where $\bm{w}\in\mathbb{R}^{d}$ and $b \in \mathbb{R}$
are trainable parameters, $g_i\in (0, 1)$, and $\sigma$ is the sigmoid function.
% \vspace{0.4cm}

\noindent
\textbf{\texttt{vector gate (vg)}}:
\begin{align}
    \bm{g}_i &= \sigma(\bm{W}\bm{v}^{(w)}_i + \bm{b})\\
    \bm{v}_i &= \bm{g}_i\odot\bm{v}^{(c)}_i + (\bm{1} - \bm{g}_i)\odot\bm{v}^{(w)}_i\label{eq:vg}
\end{align}

\noindent 
where $\bm{W} \in \mathbb{R}^{d \times d}$ and $\bm{b}
\in \mathbb{R}^{d}$ are trainable parameters, $\bm{g}_i \in {(0,
1)}^{d}$, $\sigma$ is the element-wise sigmoid function, $\odot$ is the
element-wise product for vectors, and $\bm{1} \in \mathbb{R}^{d}$ is a 
vector of ones.

The vector gate is inspired by~\citet{miyamoto2016gated}
and~\citet{yang2017words}, but is different to the former in that the gating
mechanism acts upon each dimension of the word and character-level vectors, and
different to the latter in that it does not rely on external sources of
information for calculating the gating mechanism.

Finally, note that \texttt{word only} and \texttt{char only} are special cases of
both gating mechanisms: $g_i = 0$ (scalar gate) and $\bm{g}_i = \bm{0}$ (vector
gate) correspond to \texttt{word only}; $g_i = 1$ and $\bm{g}_i =
\bm{1}$ correspond to \texttt{char only}.

% FIXME: Consider mentioning why I used these gating mechanisms: Conceptually
% simple to grasp and implement

% The rationale behind the vector gate mechanism is that some tasks might benefit
% from the added modeling complexity it offers, as shown by
% \citet{yang2017words}, while being simple enough for not needing access to
% external knowledge.

\subsection{Obtaining Sentence Representations}

To enable sentence-level classification we need to obtain a sentence
representation from the word vectors $\bm{v}_i$. We achieved this by using a
\ac{BiLSTM} with max pooling, which was shown to be a good universal sentence
encoding mechanism~\citep{conneau2017supervised}.

Let $\bm{x} = w_1, \ldots, w_n$, be an input sentence and $\bm{V} = [\bm{v}_1,
\ldots, \bm{v}_n]$ its matrix representation, where each $\bm{v}_i$ was obtained
by one of the methods described in \cref{subsec:methods}. $\bm{S} = [\bm{s}_1,
\ldots, \bm{s}_n]$ is the context-enriched matrix representation of $\bm{x}$
obtained by feeding $\bm{V}$ to a \ac{BiLSTM} of output dimension
$d_s$\footnote{$\bm{s}_i = [\overrightarrow{\bm{s}_i};
\overleftarrow{\bm{s}_i}]$ for each $i$, and both $\overrightarrow{\bm{s}_i}$
and $\overleftarrow{\bm{s}_i} \in \mathbb{R}^{\frac{d_s}{2}}$.}. Lastly,
$\bm{s} \in \mathbb{R}^{d_s}$ is the final sentence representation of $\bm{x}$
obtained by max-pooling $\bm{S}$ along the sequence dimension.

Finally, we initialized the word representations $\bm{v}^{(w)}_i$ using GloVe
embeddings \citep{pennington2014glove}, and fine-tuned them during training.
Refer to \cref{app:hyperparams} for details on the other hyperparameters
we used.

% models are implemented in the same
% fashion as the BiLSTM with max pooling described
% in~\citet{conneau2017supervised}. Refer to Appendix \ref{app:hyperparams} for
% details on the hyperparameters we used.

\section{Experiments}\label{sec:experiments}
\subsection{Experimental Setup}\label{subsec:experimental-setup}

We trained our models for solving the Natural Language Inference (NLI) task in
two datasets, SNLI~\citep{bowman2015large} and MultiNLI~\citep{williams2018broad},
and validated them in each corresponding development set (including the matched
and mismatched development sets of MultiNLI).

For each dataset-method combination we trained 7 models initialized with
different random seeds, and saved each when it reached its best validation
accuracy\footnote{We found that models validated on the matched development set
of MultiNLI, rather than the mismatched, yielded best results, although the
differences were not statistically significant.}. We then evaluated the quality
of each trained model's word representations $\bm{v}_i$ in 10 word similarity
tasks, using the system created by
\citet{jastrzebski2017evaluate}\footnote{{\tiny
\url{https://github.com/kudkudak/word-embeddings-benchmarks/tree/8fd0489}}}.

Finally, we fed these obtained word vectors to a \ac{BiLSTM} with max-pooling
and evaluated the final sentence representations in 11 downstream transfer
tasks~\citep{conneau2017supervised, subramanian2018learning}.

\begin{table*}[t!]
\scriptsize
% \begin{adjustwidth}{-1cm}
\centering

\begin{tabular}{llcccccccccc}
\toprule
     &              & \tbf{MEN}        & \tbf{MTurk287}      & \tbf{MTurk771}      & \tbf{RG65}       & \tbf{RW}         & \tbf{SimLex999}      & \tbf{SimVerb3500}     & \tbf{WS353}      & \tbf{WS353R}     & \tbf{WS353S}     \\
\midrule
SNLI & \texttt{w}   & 71.78            & 35.40            & 49.05            & 61.80            & 18.43            & 19.17            & 10.32            & 39.27            & 28.01            & 53.42            \\
     & \texttt{c}   & 9.85             & -5.65            & 0.82             & -5.28            & 17.81            & 0.86             & 2.76             & -2.20            & 0.20             & -3.87            \\
     & \texttt{cat} & 71.91            & \tbf{35.52}      & 48.84            & 62.12            & 18.46            & 19.10            & 10.21            & 39.35            & 28.16            & 53.40            \\
     & \texttt{sg}  & 70.49            & 34.49            & 46.15            & 59.75            & 18.24            & 17.20            & 8.73             & 35.86            & 23.48            & 50.83            \\
     & \texttt{vg}  & \ul{\tbf{80.00}} & 32.54            & \tbf{62.09}      & \tbf{68.90}      & \tbf{20.76}      & \tbf{37.70}      & \tbf{20.45}      & \tbf{54.72}      & \tbf{47.24}      & \tbf{65.60}      \\
\midrule
MNLI & \texttt{w}   & 68.76            & 50.15            & 68.81            & 65.83            & 18.43            & 42.21            & 25.18            & 61.10            & 58.21            & 70.17            \\
     & \texttt{c}   & 4.84             & 0.06             & 1.95             & -0.06            & 12.18            & 3.01             & 1.52             & -4.68            & -3.63            & -3.65            \\
     & \texttt{cat} & 68.77            & 50.40            & 68.77            & 65.92            & 18.35            & 42.22            & 25.12            & 61.15            & 58.26            & 70.21            \\
     & \texttt{sg}  & 67.66            & 49.58            & 68.29            & 64.84            & 18.36            & 41.81            & 24.57            & 60.13            & 57.09            & 69.41            \\
     & \texttt{vg}  & \tbf{76.69}      & \ul{\tbf{56.06}} & \ul{\tbf{70.13}} & \ul{\tbf{69.00}} & \ul{\tbf{25.35}} & \ul{\tbf{48.40}} & \ul{\tbf{35.12}} & \ul{\tbf{68.91}} & \ul{\tbf{64.70}} & \ul{\tbf{77.23}} \\
\bottomrule
\end{tabular}

\caption{Word-level evaluation results. Each value corresponds to
    average Pearson correlation of 7 identical models initialized with different
    random seeds. Correlations were scaled to the $[-100; 100]$ range for easier
    reading. \tbf{Bold} values represent the best method per training dataset,
    per task; \ul{\tbf{underlined}} values represent the best-performing method
    per task, independent of training dataset. For each task and dataset, every
    best-performing method was significantly different to other methods
    ($p<0.05$), except for \texttt{w} trained in SNLI at the MTurk287 task.
    Statistical significance was obtained with a two-sided Welch's t-test for
    two independent samples without assuming equal
variance~\citep{welch1947generalization}.}%

\label{table:word_level_results}

% \end{adjustwidth}
\end{table*}

\subsection{Datasets}\label{subsec:datasets}

\textbf{Word-level Semantic Similarity} A desirable property of vector
representations of words is that semantically similar words should have similar
vector representations. Assessing whether a set of word representations
possesses this quality is referred to as the semantic similarity task. This is
the most widely-used evaluation method for evaluating word representations,
despite its shortcomings \citep{faruqui2016problems}.

This task consists of comparing the similarity between word vectors measured by
a distance metric (usually cosine distance), with a similarity score obtained
from human judgements. High correlation between these similarities is an
indicator of good performance.

A problem with this formulation though, is that the definition of ``similarity''
often confounds the meaning of both \textit{similarity} and
\textit{relatedness}. For example, \textit{cup} and \textit{tea} are related but
dissimilar words, and this type of distinction is not always clear
\citep{agirre2009study, hill2015simlex}.

To face the previous problem, we tested our methods in a wide variety of
datasets, including some that explicitly model relatedness (WS353R), some that
explicitly consider similarity (WS353S, SimLex999, SimVerb3500), and some where
the distinction is not clear (MEN, MTurk287, MTurk771, RG, WS353). We also
included the RareWords (RW) dataset for evaluating the quality of rare word
representations. See \cref{appendix:datasets} for a more complete description of
the datasets we used.

\textbf{Sentence-level Evaluation Tasks} Unlike word-level representations,
there is no consensus on the desirable properties sentence representations
should have. In response to this, \citet{conneau2017supervised} created
SentEval\footnote{{\tiny
\url{https://github.com/facebookresearch/SentEval/tree/906b34a}}}, a sentence
representation evaluation benchmark designed for assessing how well sentence
representations perform in various downstream tasks \citep{conneau2018senteval}.

Some of the datasets included in SentEval correspond to sentiment classification
(CR, MPQA, MR, SST2, and SST5), subjectivity classification (SUBJ),
question-type classification (TREC), recognizing textual entailment (SICK E),
estimating semantic relatedness (SICK R), and measuring textual semantic
similarity (STS16, STSB). The datasets are described by
\citet{conneau2017supervised}, and we provide pointers to their original sources
in the appendix \cref{table:sentence-eval-datasets}.

To evaluate these sentence representations SentEval trained a linear model on
top of them, and evaluated their performance in the validation sets accompanying
each dataset. The only exception was the STS16 task, in which our representations
were evaluated directly.

% The subset of tasks included in SentEval in which we tested our generated
% sentence representations correspond to classification (CR, MPQA, MR, SST2, SST5,
% SUBJ, TREC), entailment (SICK E), semantic relatedness (SICK R), and semantic
% textual similarity (STS16, STS B), and are described

% \section{Results and Discussion}

\section{Word-level Evaluation}
\label{sec:word_level_eval}

\subsection{Word Similarity}
\label{subsec:word-similarity-eval}

\Cref{table:word_level_results} shows the quality of word representations
in terms of the correlation between word similarity scores obtained by the
proposed models and word similarity scores defined by humans.

First, we can see that for each task, \texttt{character only} models had
significantly worse performance than every other model trained on the same
dataset. The most likely explanation for this is that these models are the only
ones that need to learn word representations from scratch, since they have no
access to the global semantic knowledge encoded by the GloVe embeddings.

Further, \tbf{bold} results show the overall trend that \texttt{vector gates}
outperformed the other methods regardless of training dataset. This implies that
learning how to combine character and word-level representations at the
dimension level produces word vector representations that capture a notion of
word similarity and relatedness that is closer to that of humans.

Additionally, results from the MNLI row in general, and \tbf{\ul{underlined}}
results in particular, show that training on MultiNLI produces word
representations better at capturing word similarity. This is probably due to
MultiNLI data being richer than that of SNLI\@. Indeed, MultiNLI data was gathered
from various sources (novels, reports, letters, and telephone conversations,
among others), rather than the single image captions dataset from which SNLI was
created.

Exceptions to the previous rule are models evaluated in MEN and RW\@. The former
case can be explained by the MEN dataset\footnote{\scriptsize
\url{https://staff.fni.uva.nl/e.bruni/MEN}} containing only words that appear as
image labels in the ESP-Game\footnote{\scriptsize
\url{http://www.cs.cmu.edu/~biglou/resources/}} and
MIRFLICKR-1M\footnote{\scriptsize \url{http://press.liacs.nl/mirflickr/}} image
datasets \citep{bruni2014men}, and therefore having data that is more closely
distributed to SNLI than to MultiNLI.

More notably, in the RareWords dataset \citep{luong2013better}, the
\texttt{word only}, \texttt{concat}, and \texttt{scalar gate} methods performed
equally, despite having been trained in different datasets ($p>0.1$), and the
\texttt{char only} method performed significantly worse when trained in
MultiNLI. The \texttt{vector gate}, however, performed significantly better than
its counterpart trained in SNLI. These facts provide evidence that this method
is capable of capturing linguistic phenomena that the other methods are unable
to model.

\begin{figure*}[!htbp]
    \centering
    \includegraphics[scale=0.25]{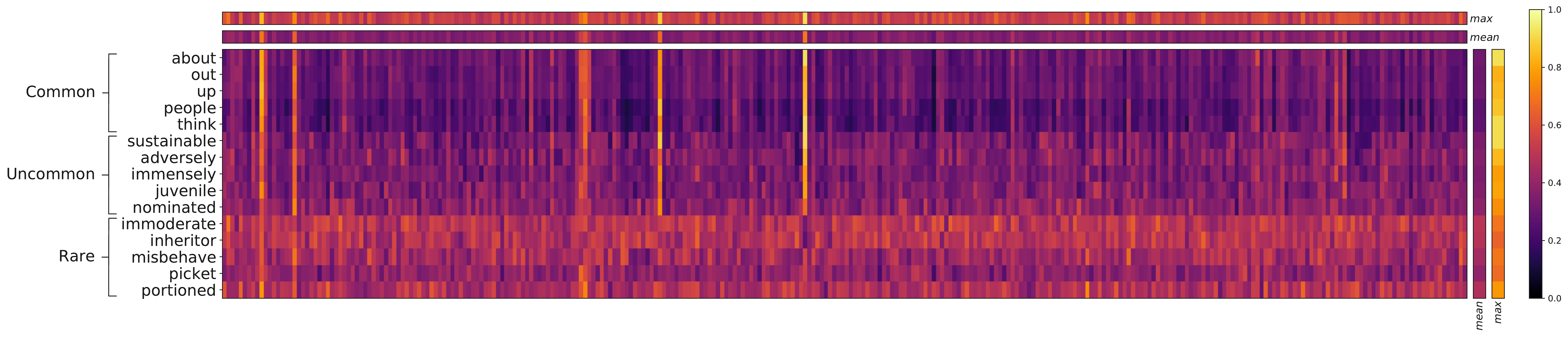}
    \caption{Visualization of gating values for $5$ common words (freq.
    $\sim20000$), $5$ uncommon words (freq. $\sim60$), and $5$ rare words (freq.
    $\sim2$), appearing in both the RW and MultiNLI datasets.}
    \label{fig:gating_viz}
\end{figure*}

% In general, the better performance of the \texttt{vector gate} across these
% tasks implies that the word representations it creates are better at modeling
% paradigmatic relationships between words \citep{saussure1915course}, that is,
% semantic relationships that hold \textit{in absentia}.

% The fact that other methods did not benefit
% from being trained in the richer MultiNLI, that the \texttt{char only} method
% performed worse when trained in MultiNLI, and that, despite the previous, the
% \texttt{vector gate} performed better, evidences that this method is capable of
% capturing linguistic phenomena that the other methods are unable to model.

\subsection{Word Frequencies and Gating Values}
\label{subsec:freqs-and-gating-values}

\begin{figure}
    \centering
    \includegraphics[width=\columnwidth]{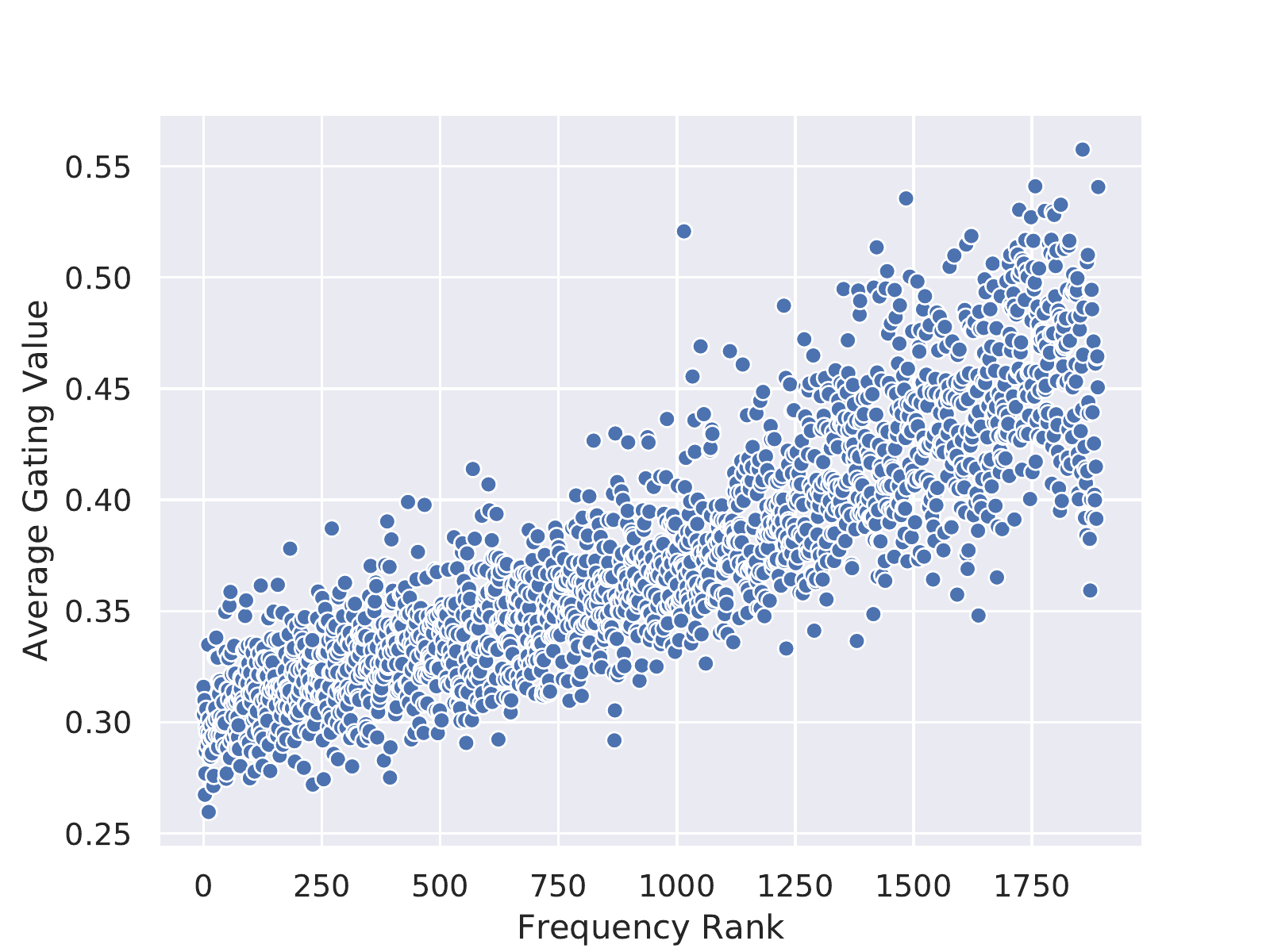}
    \caption{Average gating values for words appearing in both RW and MultiNLI.
    Words are sorted by decreasing frequency in MultiNLI.}
    \label{fig:freq_vs_gate_value}
\end{figure}

\Cref{fig:gating_viz} shows that for more common words the \texttt{vector
gate} mechanism tends to favor only a few dimensions while keeping a low average
gating value across dimensions. On the other hand, values are greater and more
homogeneous across dimensions in rarer words. Further,
\cref{fig:freq_vs_gate_value} shows this mechanism assigns, on average, a
greater gating value to less frequent words, confirming the findings by
\citet{miyamoto2016gated}, and \citet{yang2017words}.

In other words, the less frequent the word, the more this mechanism allows the
character-level representation to influence the final word representation, as
shown by \cref{eq:vg}. A possible interpretation of this result is that
exploiting character information becomes increasingly necessary as word-level
representations' quality decrease.

Another observable trend in both figures is that gating values tend to be low on
average. Indeed, it is possible to see in \cref{fig:freq_vs_gate_value}
that the average gating values range from $0.26$ to $0.56$. This result
corroborates the findings by \citet{miyamoto2016gated}, stating that setting
$g=0.25$ in \cref{eq:scalar-gate}, was better than setting it to higher
values.

In summary, the gating mechanisms learn how to compensate the lack of
expressivity of underrepresented words by selectively combining their
representations with those of characters.

\section{Sentence-level Evaluation}

\Cref{table:sent_level_results} shows the impact that different methods for
combining character and word-level word representations have in the quality of
the sentence representations produced by our models.

% The NLI column corresponds to the best validation accuracy obtained during each
% model's training procedure. It shows that models trained on the SNLI dataset do
% not benefit from using character information. Indeed, both word-only
% (\texttt{w}) and concat (\texttt{cat}) methods perform roughly equally. This
% might be due to both training and development data distributions being similar
% for this dataset, and therefore not needing the extra information for rare or
% unknown words provided by the character-level granularity. On the other hand,
% models trained on MultiNLI did benefit from using characters both when
% evaluating in the matched and mismatched datasets.

We can observe the same trend mentioned in \cref{subsec:word-similarity-eval},
and highlighted by the difference between \tbf{bold} values, that models trained
in MultiNLI performed better than those trained in SNLI at a statistically
significant level, confirming the findings of \citet{conneau2017supervised}.
In other words, training sentence encoders on MultiNLI yields more general
sentence representations than doing so on SNLI.

% This confirms
% the findings of  that training on MultiNLI yields
% more general sentence representations than training on SNLI only

% \tbf{\ul{Underlined}} results show that most tasks ($9$~/~$11$), benefit from
% models being trained on MultiNLI rather than SNLI, at a statistically
% significant level. These differences confirm the findings of
% \citet{conneau2017supervised} that training on MultiNLI yields more general
% sentence representations than training on SNLI only.

The two exceptions to the previous trend, SICKE and SICKR, benefited more from
models trained on SNLI\@. We hypothesize this is again due to both SNLI
and SICK~\citep{marelli2014sick} having similar data distributions\footnote{SICK
    was created from Flickr-8k~\citep{rashtchian2010flickr8k}, and SNLI from its
expanded version: Flickr30k~\citep{young2014flickr30k}.}.

\begin{table*}[t!]
\scriptsize
% \begin{adjustwidth}{-1cm}
\centering

\begin{tabular}{llcccccccccccccccccccc}
\toprule

     &                    & \multicolumn{7}{c}{Classification} & \multicolumn{1}{c}{Entailment} & \multicolumn{1}{c}{Relatedness} & \multicolumn{2}{c}{Semantic Textual Similarity} \\
\cmidrule(lr){3-9}
\cmidrule(lr){10-10}
\cmidrule(lr){11-11}
\cmidrule(lr){12-13}
     &              & \tbf{CR}               & \tbf{MPQA}       & \tbf{MR}         & \tbf{SST2}       & \tbf{SST5}       & \tbf{SUBJ}             & \tbf{TREC}       & \tbf{SICKE}      & \tbf{SICKR$^\dag$} & \tbf{STS16$^\dag$} & \tbf{STSB$^\dag$} \\
\midrule
SNLI & \texttt{w}   & 80.50                  & 84.59            & 74.18            & 78.86            & 42.33            & \tbf{90.38}            & \tbf{86.83}      & 86.37            & 88.52              & 59.90$^{*}$        & 71.29$^{*}$       \\
     & \texttt{c}   & 74.90$^{*}$            & 78.86$^{*}$      & 65.93$^{*}$      & 69.42$^{*}$      & 35.56$^{*}$      & 82.97$^{*}$            & 83.31$^{*}$      & 84.13$^{*}$      & 83.89$^{*}$        & 59.33$^{*}$        & 67.20$^{*}$       \\
     & \texttt{cat} & 80.44                  & 84.66            & 74.31            & 78.37            & 41.34$^{*}$      & 90.28                  & 85.80$^{*}$      & \ul{\tbf{86.40}} & 88.44              & 59.90$^{*}$        & 71.24$^{*}$       \\
     & \texttt{sg}  & \tbf{80.59}            & 84.60            & \tbf{74.49}      & \tbf{79.04}      & 41.63$^{*}$      & 90.16                  & 86.00            & 86.10$^{*}$      & \ul{\tbf{88.57}}   & 60.05$^{*}$        & 71.34$^{*}$       \\
     & \texttt{vg}  & 80.42                  & \tbf{84.66}      & 74.26            & 78.87            & \tbf{42.38}      & 90.07                  & 85.97            & 85.67            & 88.31$^{*}$        & \tbf{60.92}        & \tbf{71.99}       \\
\midrule
MNLI & \texttt{w}   & 83.80                  & \ul{\tbf{89.13}} & 79.05            & 83.38            & 45.21            & 91.79                  & 89.23            & 84.92            & 86.33              & 66.08              & 71.96$^{*}$       \\
     & \texttt{c}   & 70.23$^{*}$            & 72.19$^{*}$      & 62.83$^{*}$      & 64.55$^{*}$      & 32.47$^{*}$      & 79.49$^{*}$            & 74.74$^{*}$      & 81.53$^{*}$      & 75.92$^{*}$        & 51.47$^{*}$        & 61.74$^{*}$       \\
     & \texttt{cat} & \ul{\tbf{83.96}}       & 89.12            & \ul{\tbf{79.23}} & 83.70            & 45.08$^{*}$      & \ul{\tbf{91.92}}       & \ul{\tbf{90.03}} & \tbf{85.06}      & 86.45              & \ul{\tbf{66.17}}   & 71.82$^{*}$       \\
     & \texttt{sg}  & 83.88                  & 89.06            & 79.22            & 83.71            & 45.26            & 91.66$^{*}$            & 88.83$^{*}$      & 84.96            & 86.40              & 65.49$^{*}$        & 71.87$^{*}$       \\
     & \texttt{vg}  & 83.45$^{*}$            & 89.05            & 79.13            & \ul{\tbf{83.87}} & \ul{\tbf{45.88}} & 91.55$^{*}$            & 89.49            & 84.82            & \tbf{86.50}        & 65.75              & \ul{\tbf{72.82}}  \\
\bottomrule

\end{tabular}

\caption{Experimental results. Each value shown in the table is the
    average result of 7 identical models initialized with different random
    seeds. Values represent accuracy (\%) unless indicated by $\dag$, in which
    case they represent Pearson correlation scaled to the range $[-100, 100]$
    for easier reading. \tbf{Bold} values represent the best method per training
    dataset, per task; \ul{\tbf{underlined}} values represent the
    best-performing method per task, independent of training dataset. Values
    marked with an asterisk ($^{*}$) are significantly different to the average
    performance of the best model trained on the same dataset ($p < 0.05$).
    Results for every best-performing method trained on one dataset are
significantly different to the best-performing method trained on the other.
Statistical significance was obtained in the same way as described in
\cref{table:word_level_results}.}%

\label{table:sent_level_results}

% \end{adjustwidth}
\end{table*}

Additionally, there was no method that significantly outperformed the
\texttt{word only} baseline in classification tasks. This means that the added
expressivity offered by explicitly modeling characters, be it through
concatenation or gating, was not significantly better than simply fine-tuning
the pre-trained GloVe embeddings for this type of task. We hypothesize this is
due to the conflation of two effects. First, the fact that morphological
processes might not encode important information for solving these tasks; and
second, that SNLI and MultiNLI belong to domains that are too dissimilar to the
domains in which the sentence representations are being tested.

On the other hand, the \texttt{vector gate} significantly outperformed every
other method in the STSB task when trained in both datasets, and in the STS16
task when trained in SNLI. This again hints at this method being capable of
modeling phenomena at the word level, resulting in improved semantic
representations at the sentence level.

\section{Relationship Between Word- and Sentence-level Evaluation Tasks}

It is clear that the better performance the \texttt{vector gate} had in word
similarity tasks did not translate into overall better performance in downstream
tasks. This confirms previous findings indicating that intrinsic word evaluation
metrics are not good predictors of downstream
performance~\citep{tsvetkov2015evaluation, chiu2016instrinsic,
faruqui2016problems, gladkova2016intrinsic}.

\begin{figure*}[!htbp]
    \centering
    \begin{subfigure}{0.495\textwidth}
        \includegraphics[width=\columnwidth]{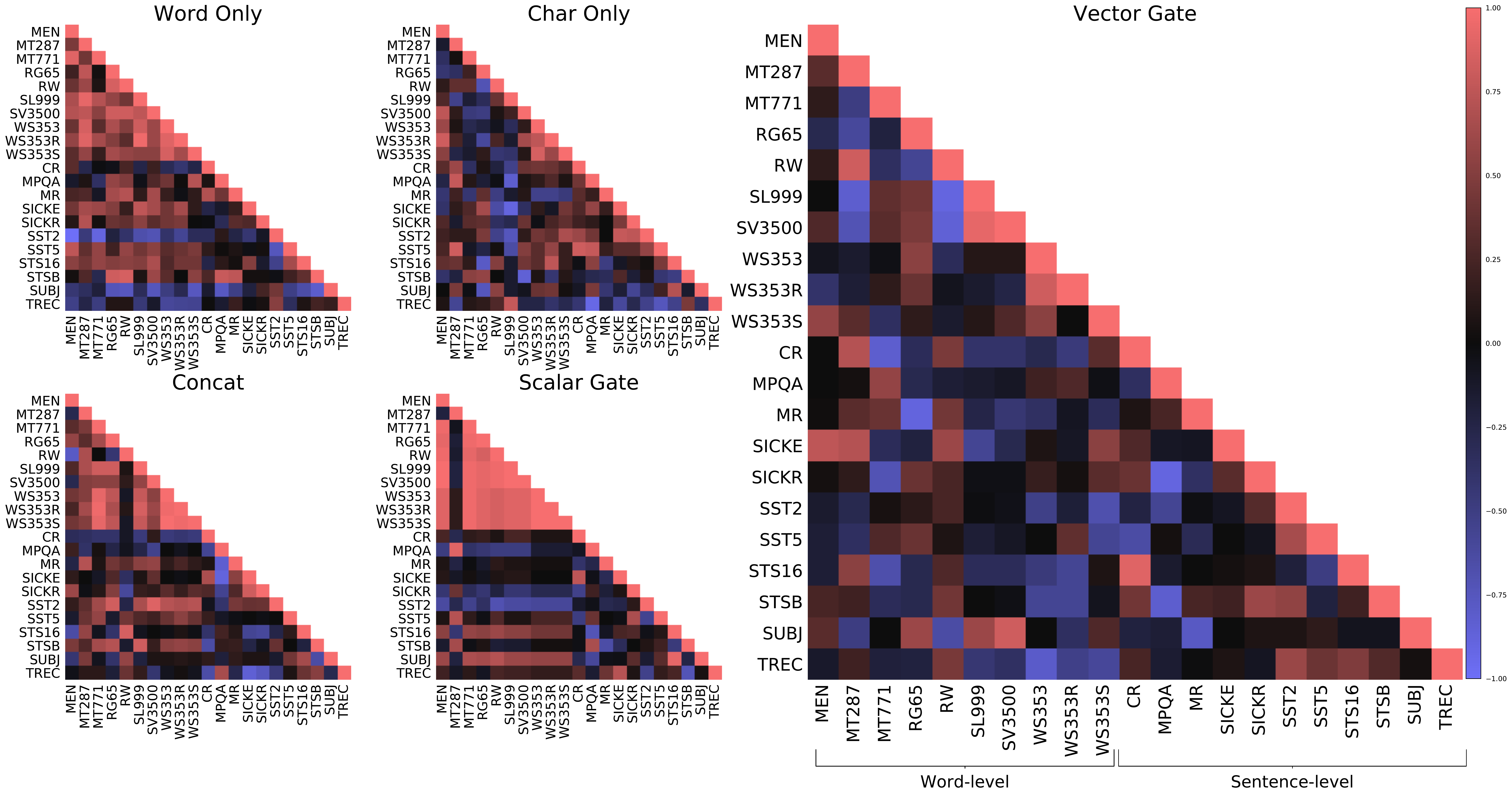}
        \caption{Models trained in SNLI.}
        \label{subfig:snli-correlations}
    \end{subfigure}
    \begin{subfigure}{0.495\textwidth}
        \includegraphics[width=\columnwidth]{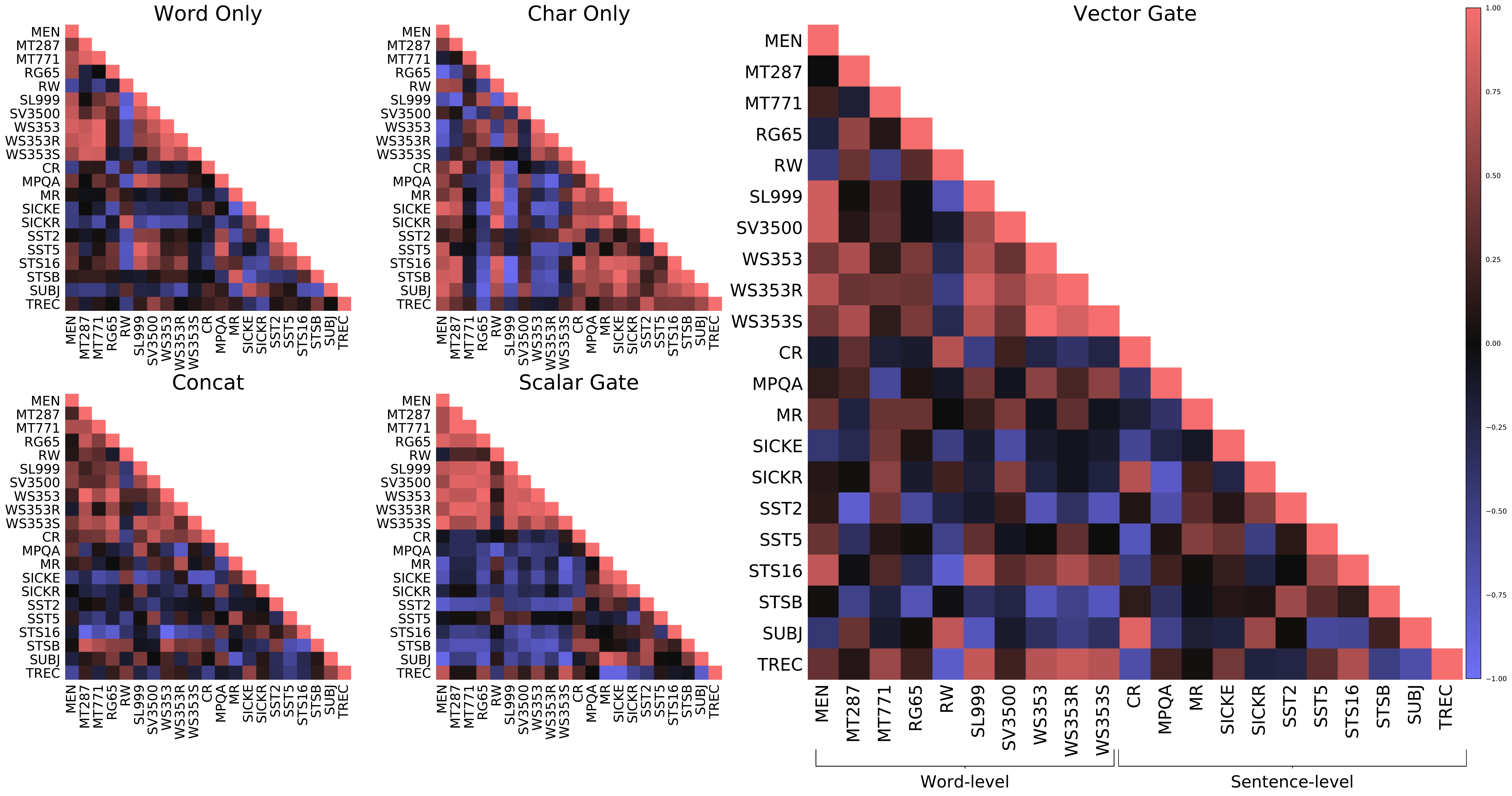}
        \caption{Models trained in MultiNLI.}
        \label{subfig:mnli-correlations}
    \end{subfigure}
    \caption{Spearman correlation between performances in word and sentence
        level evaluation tasks.}\label{fig:correlations}
\end{figure*}

\Cref{subfig:mnli-correlations} shows that the word representations created by
the \texttt{vector gate} trained in MultiNLI had positively-correlated results
within several word-similarity tasks. This hints at the generality of the word
representations created by this method when modeling similarity and relatedness.

% This hints at the generality of these
% when capturing paradigmatic

% In other words, This means that the representations created by
% this method were general enough to perform well in a variety of tasks evaluating
% similarity, relatedness and a conflation of both.

However, the same cannot be said about sentence-level evaluation performance;
there is no clear correlation between word similarity tasks and
sentence-evaluation tasks. This is clearly illustrated by performance in the
STSBenchmark, the only in which the \texttt{vector gate} was significantly
superior, not being correlated with performance in any word-similarity dataset.
This can be interpreted simply as word-level representations capturing
word-similarity not being a sufficient condition for good performance in
sentence-level tasks.

In general, \cref{fig:correlations} shows that there are no general
correlation effects spanning both training datasets and combination mechanisms.
For example, \cref{subfig:snli-correlations} shows that, for both
\texttt{word-only} and \texttt{concat} models trained in SNLI, performance in
word similarity tasks correlates positively with performance in most sentence
evaluation tasks, however, this does not happen as clearly for the same models
trained in MultiNLI (\cref{subfig:mnli-correlations}).

\section{Related Work}
\label{sec:related-work}

\subsection{Gating Mechanisms for Combining Characters and Word Representations}

To the best of our knowledge, there are only two recent works that specifically
study how to combine word and subword-level vector representations.

\citet{miyamoto2016gated} propose to use a trainable scalar gating mechanism
capable of learning a weighting scheme for combining character-level and
word-level representations. They compared their proposed method to manually
weighting both levels; using characters only; words only; or their
concatenation. They found that in some datasets a specific manual weighting
scheme performed better, while in others the learned scalar gate did.

\citet{yang2017words} further expand the gating concept by making the
mechanism work at a finer-grained level, learning how to weight each vector's
dimensions independently, conditioned on external word-level features such as
part-of-speech and named-entity tags. Similarly, they compared their proposed
mechanism to using words only, characters only, and a concatenation of both,
with and without external features. They found that their vector gate performed
better than the other methods in all the reported tasks, and beat the state of
the art in two reading comprehension tasks.

Both works showed that the gating mechanisms assigned greater importance to
character-level representations in rare words, and to word-level representations
in common ones, reaffirming the previous findings that subword structures in
general, and characters in particular, are beneficial for modeling uncommon
words.

\subsection{Sentence Representation Learning}

The problem of representing sentences as fixed-length vectors has been widely
studied.
% \citep{kiros2015skipthought,hill2016learning,radford2017learning,
% conneau2017supervised,subramanian2018learning}

% The problem has been formulated as both requiring and not requiring labeled
% data; often called supervised and unsupervised respectively.

\citet{zhao2015self-adaptive} suggested a self-adaptive hierarchical model that
gradually composes words into intermediate phrase representations, and
adaptively selects specific hierarchical levels for specific tasks.
\citet{kiros2015skipthought} proposed an encoder-decoder model trained by
attempting to reconstruct the surrounding sentences of an encoded passage, in a
fashion similar to Skip-gram \citep{mikolov2013distributed}.
\citet{hill2016learning} overcame the previous model's need for ordered training
sentences by using autoencoders for creating the sentence representations.
\citet{jernite2017discourse} implemented a model simpler and faster to train
than the previous two, while having competitive performance. Similar to
\citet{kiros2015skipthought}, \citet{gan2017learning} suggested predicting
future sentences with a hierarchical CNN-LSTM encoder.

% proposed training an encoder on unlabeled
% data using several objective functions capturing different aspects of
% discourse coherence, resulting in a model with competitive results but trainable
% in significantly less time.

\citet{conneau2017supervised} trained several sentence encoding architectures on
a combination of the SNLI and MultiNLI datasets, and showed that a \ac{BiLSTM}
with max-pooling was the best at producing highly transferable sentence
representations. More recently, \citet{subramanian2018learning} empirically
showed that sentence representations created in a multi-task setting
\citep{collobert2008unified}, performed increasingly better the more tasks they
were trained in. \citet{zhang2018learning} proposed using an autoencoder that
relies on multi-head self-attention over the concatenation of the max and mean
pooled encoder outputs for producing sentence representations. Finally,
\citet{wieting2019no} show that modern sentence embedding methods are not vastly
superior to random methods.

The works mentioned so far usually evaluate the quality of the produced sentence
representations in sentence-level downstream tasks. Common benchmarks grouping
these kind of tasks include SentEval \citep{conneau2018senteval}, and GLUE
\citep{wang2019glue}. Another trend, however, is to \emph{probe} sentence
representations to understand what linguistic phenomena they encode
\citep{linzen2016assessing, adi2017fine, conneau2018what, perone2018evaluation,
zhu2018exploring}.

\subsection{General Feature-wise Transformations}

\citet{dumoulin2018feature-wise} provide a review on feature-wise
transformation methods, of which the mechanisms presented in this paper form a
part of. In a few words, the $g$ parameter, in both \texttt{scalar gate} and
\texttt{vector gate} mechanisms, can be understood as a \emph{scaling parameter}
limited to the $(0,1)$ range and conditioned on word representations, whereas
adding the scaled $\bm{v}_i^{(c)}$ and $\bm{v}_i^{(w)}$ representations can be
seen as \emph{biasing} word representations conditioned on character
representations.

The previous review extends the work by \citet{perez2018film}, which describes
the Feature-wise Linear Modulation (FiLM) framework as a generalization of
Conditional Normalization methods, and apply it in visual reasoning tasks. Some
of the reported findings are that, in general, scaling has greater impact than
biasing, and that in a setting similar to the \texttt{scalar gate}, limiting the
scaling parameter to $(0,1)$ hurt performance. Future decisions involving the
design of mechanisms for combining character and word-level representations
should be informed by these insights.

\section{Conclusions}

We presented an empirical study showing the effect that different ways of
combining character and word representations has in word-level and
sentence-level evaluation tasks.

We showed that a vector gate performed consistently better across a variety of
word similarity and relatedness tasks. Additionally, despite showing
inconsistent results in sentence evaluation tasks, it performed significantly
better than the other methods in semantic similarity tasks.

We further showed through this mechanism, that learning character-level
representations is always beneficial, and becomes increasingly so with less
common words.

In the future it would be interesting to study how the choice of mechanism for
combining subword and word representations affects the more recent
language-model-based pretraining methods such as ELMo \citep{peters2018elmo},
GPT \citep{radford2018improving, radford2019language} and BERT
\citep{devlin2018bert}.

\section*{Acknowledgements}

Thanks to Edison Marrese-Taylor and Pablo Loyola for their feedback on early
versions of this manuscript. We also gratefully acknowledge the support of the
NVIDIA Corporation with the donation of one of the GPUs used for this research.
Jorge A. Balazs is partially supported by the Japanese Government MEXT
Scholarship.

\nocite{paszke2017automatic}
\nocite{oliphant2015numpy}
\nocite{jones2001scipy}
\nocite{mckinney2010pandas}
\nocite{waskom2018seaborn}
\nocite{hunter2007matplotlib}
\nocite{rossum1995python}

\let\interlinepenaltyold=\interlinepenalty
\interlinepenalty=10000
\bibliography{bibliography}

\begin{thebibliography}{78}
\expandafter\ifx\csname natexlab\endcsname\relax\def\natexlab#1{#1}\fi

\bibitem[{Adi et~al.(2017)Adi, Kermany, Belinkov, Lavi, and
  Goldberg}]{adi2017fine}
Yossi Adi, Einat Kermany, Yonatan Belinkov, Ofer Lavi, and Yoav Goldberg. 2017.
\newblock \href {https://openreview.net/pdf?id=BJh6Ztuxl} {{Fine-grained
  Analysis of Sentence Embeddings Using Auxiliary Prediction Tasks}}.
\newblock In \emph{Proceedings of the 5th International Conference on Learning
  Representations (ICLR)}, Toulon, France.

\bibitem[{Agirre et~al.(2009)Agirre, Alfonseca, Hall, Kravalova, Pasca, and
  Soroa}]{agirre2009study}
Eneko Agirre, Enrique Alfonseca, Keith Hall, Jana Kravalova, Marius Pasca, and
  Aitor Soroa. 2009.
\newblock \href {http://aclweb.org/anthology/N09-1003} {{A Study on Similarity
  and Relatedness Using Distributional and WordNet-based Approaches}}.
\newblock In \emph{Proceedings of Human Language Technologies: The 2009 Annual
  Conference of the North American Chapter of the Association for Computational
  Linguistics}, pages 19--27, Boulder, Colorado. Association for Computational
  Linguistics.

\bibitem[{Agirre et~al.(2016)Agirre, Banea, Cer, Diab, Gonzalez-Agirre,
  Mihalcea, Rigau, and Wiebe}]{agirre2016semeval}
Eneko Agirre, Carmen Banea, Daniel Cer, Mona Diab, Aitor Gonzalez-Agirre, Rada
  Mihalcea, German Rigau, and Janyce Wiebe. 2016.
\newblock \href {http://www.aclweb.org/anthology/S16-1081} {{SemEval-2016 Task
  1: Semantic Textual Similarity, Monolingual and Cross-Lingual Evaluation}}.
\newblock In \emph{Proceedings of the 10th International Workshop on Semantic
  Evaluation (SemEval-2016)}, pages 497--511, San Diego, California.
  Association for Computational Linguistics.

\bibitem[{Arora et~al.(2017)Arora, Liang, and Ma}]{arora2017simple}
Sanjeev Arora, Yingyu Liang, and Tengyu Ma. 2017.
\newblock \href {https://openreview.net/pdf?id=SyK00v5xx} {{A Simple but
  Tough-to-Beat Baseline for Sentence Embeddings}}.
\newblock In \emph{International Conference on Learning Representations}.

\bibitem[{Avraham and Goldberg(2017)}]{avraham2017interplay}
Oded Avraham and Yoav Goldberg. 2017.
\newblock \href {https://arxiv.org/abs/1704.01938} {{The Interplay of Semantics
  and Morphology in Word Embeddings}}.
\newblock \emph{arXiv preprint arXiv:1704.01938}.

\bibitem[{Bojanowski et~al.(2017)Bojanowski, Grave, Joulin, and
  Mikolov}]{bojanowski2017enriching}
Piotr Bojanowski, Edouard Grave, Armand Joulin, and Tomas Mikolov. 2017.
\newblock \href {https://transacl.org/ojs/index.php/tacl/article/view/999}
  {{Enriching Word Vectors with Subword Information}}.
\newblock \emph{Transactions of the Association for Computational Linguistics},
  5:135--146.

\bibitem[{Botha and Blunsom(2014)}]{botha2014compositional}
Jan Botha and Phil Blunsom. 2014.
\newblock \href {http://proceedings.mlr.press/v32/botha14.html} {{Compositional
  Morphology for Word Representations and Language Modelling}}.
\newblock In \emph{Proceedings of the 31st International Conference on Machine
  Learning}, volume~32 of \emph{Proceedings of Machine Learning Research},
  pages 1899--1907, Bejing, China. PMLR.

\bibitem[{Bowman et~al.(2015)Bowman, Angeli, Potts, and
  Manning}]{bowman2015large}
Samuel~R. Bowman, Gabor Angeli, Christopher Potts, and Christopher~D. Manning.
  2015.
\newblock \href {http://aclweb.org/anthology/D15-1075} {{A Large Annotated
  Corpus for Learning Natural Language Inference}}.
\newblock In \emph{Proceedings of the 2015 Conference on Empirical Methods in
  Natural Language Processing}, pages 632--642, Lisbon, Portugal. Association
  for Computational Linguistics.

\bibitem[{Bruni et~al.(2014)Bruni, Tran, and Baroni}]{bruni2014men}
Elia Bruni, Nam-Khanh Tran, and Marco Baroni. 2014.
\newblock \href {https://www.jair.org/index.php/jair/article/view/10857/25905}
  {{Multimodal Distributional Semantics}}.
\newblock \emph{Journal of Artificial Intelligence Research}, 49:1--47.

\bibitem[{Cer et~al.(2017)Cer, Diab, Agirre, Lopez-Gazpio, and
  Specia}]{cer2017semeval}
Daniel Cer, Mona Diab, Eneko Agirre, Inigo Lopez-Gazpio, and Lucia Specia.
  2017.
\newblock \href {https://doi.org/10.18653/v1/S17-2001} {{SemEval-2017 Task 1:
  Semantic Textual Similarity Multilingual and Crosslingual Focused
  Evaluation}}.
\newblock In \emph{Proceedings of the 11th International Workshop on Semantic
  Evaluation (SemEval-2017)}, pages 1--14, Vancouver, Canada. Association for
  Computational Linguistics.

\bibitem[{Chiu et~al.(2016)Chiu, Korhonen, and Pyysalo}]{chiu2016instrinsic}
Billy Chiu, Anna Korhonen, and Sampo Pyysalo. 2016.
\newblock \href {http://anthology.aclweb.org/W16-2501} {{Intrinsic Evaluation
  of Word Vectors Fails to Predict Extrinsic Performance}}.
\newblock In \emph{Proceedings of the 1st Workshop on Evaluating Vector-Space
  Representations for NLP}, pages 1--6, Berlin, Germany. Association for
  Computational Linguistics.

\bibitem[{Collobert and Weston(2008)}]{collobert2008unified}
Ronan Collobert and Jason Weston. 2008.
\newblock \href {https://icml.cc/Conferences/2008/papers/391.pdf} {{A Unified
  Architecture for Natural Language Processing: Deep Neural Networks with
  Multitask Learning}}.
\newblock In \emph{Proceedings of the 25th Annual International Conference on
  Machine Learning (ICML 2008)}, pages 160--167, Helsinki, Finland.

\bibitem[{Conneau and Kiela(2018)}]{conneau2018senteval}
Alexis Conneau and Douwe Kiela. 2018.
\newblock \href {http://aclweb.org/anthology/L18-1269} {{SentEval: An
  Evaluation Toolkit for Universal Sentence Representations}}.
\newblock In \emph{Proceedings of the Eleventh International Conference on
  Language Resources and Evaluation (LREC-2018)}, Miyazaki, Japan. European
  Language Resource Association.

\bibitem[{Conneau et~al.(2017)Conneau, Kiela, Schwenk, Barrault, and
  Bordes}]{conneau2017supervised}
Alexis Conneau, Douwe Kiela, Holger Schwenk, Lo\"{i}c Barrault, and Antoine
  Bordes. 2017.
\newblock \href {https://www.aclweb.org/anthology/D17-1070} {{Supervised
  Learning of Universal Sentence Representations from Natural Language
  Inference Data}}.
\newblock In \emph{Proceedings of the 2017 Conference on Empirical Methods in
  Natural Language Processing}, pages 670--680, Copenhagen, Denmark.
  Association for Computational Linguistics.

\bibitem[{Conneau et~al.(2018)Conneau, Kruszewski, Lample, Barrault, and
  Baroni}]{conneau2018what}
Alexis Conneau, Germ{\'a}n Kruszewski, Guillaume Lample, Lo{\"\i}c Barrault,
  and Marco Baroni. 2018.
\newblock \href {https://www.aclweb.org/anthology/P18-1198} {What you can cram
  into a single {\$}{\&}!{\#}* vector: Probing sentence embeddings for
  linguistic properties}.
\newblock In \emph{Proceedings of the 56th Annual Meeting of the Association
  for Computational Linguistics (Volume 1: Long Papers)}, pages 2126--2136,
  Melbourne, Australia. Association for Computational Linguistics.

\bibitem[{Devlin et~al.(2018)Devlin, Chang, Lee, and
  Toutanova}]{devlin2018bert}
Jacob Devlin, Ming{-}Wei Chang, Kenton Lee, and Kristina Toutanova. 2018.
\newblock \href {http://arxiv.org/abs/1810.04805} {{BERT: Pre-training of Deep
  Bidirectional Transformers for Language Understanding}}.
\newblock \emph{CoRR}, abs/1810.04805.

\bibitem[{Dumoulin et~al.(2018)Dumoulin, Perez, Schucher, Strub, Vries,
  Courville, and Bengio}]{dumoulin2018feature-wise}
Vincent Dumoulin, Ethan Perez, Nathan Schucher, Florian Strub, Harm~de Vries,
  Aaron Courville, and Yoshua Bengio. 2018.
\newblock \href {https://doi.org/undefined} {{Feature-wise transformations}}.
\newblock \emph{Distill}.

\bibitem[{Faruqui et~al.(2016)Faruqui, Tsvetkov, Rastogi, and
  Dyer}]{faruqui2016problems}
Manaal Faruqui, Yulia Tsvetkov, Pushpendre Rastogi, and Chris Dyer. 2016.
\newblock \href {http://anthology.aclweb.org/W16-2506} {{Problems With
  Evaluation of Word Embeddings Using Word Similarity Tasks}}.
\newblock In \emph{Proceedings of the 1st Workshop on Evaluating Vector-Space
  Representations for NLP}, pages 30--35, Berlin, Germany. Association for
  Computational Linguistics.

\bibitem[{Fellbaum(1998)}]{fellbaum98wordnet}
Christiane Fellbaum, editor. 1998.
\newblock \emph{{WordNet: an Electronic Lexical Database}}.
\newblock MIT Press.

\bibitem[{Finkelstein et~al.(2002)Finkelstein, Gabrilovich, Matias, Rivlin,
  Solan, Wolfman, and Ruppin}]{finkelstein2002wordsim353}
Lev Finkelstein, Evgeniy Gabrilovich, Yossi Matias, Ehud Rivlin, Zach Solan,
  Gadi Wolfman, and Eytan Ruppin. 2002.
\newblock \href {https://doi.org/10.1145/503104.503110} {{Placing Search in
  Context: The Concept Revisited}}.
\newblock \emph{ACM Transactions on Information Systems}, 20(1):116--131.

\bibitem[{Gan et~al.(2017)Gan, Pu, Henao, Li, He, and Carin}]{gan2017learning}
Zhe Gan, Yunchen Pu, Ricardo Henao, Chunyuan Li, Xiaodong He, and Lawrence
  Carin. 2017.
\newblock \href {https://doi.org/10.18653/v1/D17-1254} {{Learning Generic
  Sentence Representations Using Convolutional Neural Networks}}.
\newblock In \emph{Proceedings of the 2017 Conference on Empirical Methods in
  Natural Language Processing}, pages 2390--2400, Copenhagen, Denmark.
  Association for Computational Linguistics.

\bibitem[{Gerz et~al.(2016)Gerz, Vuli\'{c}, Hill, Reichart, and
  Korhonen}]{gerz2016simverb}
Daniela Gerz, Ivan Vuli\'{c}, Felix Hill, Roi Reichart, and Anna Korhonen.
  2016.
\newblock \href {https://aclweb.org/anthology/D16-1235} {{SimVerb-3500: A
  Large-Scale Evaluation Set of Verb Similarity}}.
\newblock In \emph{Proceedings of the 2016 Conference on Empirical Methods in
  Natural Language Processing}, pages 2173--2182, Austin, Texas. Association
  for Computational Linguistics.

\bibitem[{Gladkova and Drozd(2016)}]{gladkova2016intrinsic}
Anna Gladkova and Aleksandr Drozd. 2016.
\newblock \href {http://anthology.aclweb.org/W16-2507} {{Intrinsic Evaluations
  of Word Embeddings: What Can We Do Better?}}
\newblock In \emph{Proceedings of the 1st Workshop on Evaluating Vector-Space
  Representations for NLP}, pages 36--42, Berlin, Germany. Association for
  Computational Linguistics.

\bibitem[{Graves et~al.(2013)Graves, Mohamed, and Hinton}]{graves2013speech}
Alex Graves, Abdel-rahman Mohamed, and Geoffrey Hinton. 2013.
\newblock \href {https://ieeexplore.ieee.org/document/6638947/} {{Speech
  Recognition with Deep Recurrent Neural Networks}}.
\newblock In \emph{Proceedings of the 2013 International Conference on
  Acoustics, Speech and Signal Processing (ICASSP)}, pages 6645--6649,
  Vancouver, Canada. IEEE.

\bibitem[{Graves and Schmidhuber(2005)}]{graves2005framewise}
Alex Graves and J{\"u}rgen Schmidhuber. 2005.
\newblock \href
  {https://www.sciencedirect.com/science/article/pii/S0893608005001206}
  {{Framewise Phoneme Classification with Bidirectional LSTM and Other Neural
  Network Architectures}}.
\newblock \emph{Neural Networks}, 18(5-6):602--610.

\bibitem[{Halawi et~al.(2012)Halawi, Dror, Gabrilovich, and
  Koren}]{halawi2012mturk771}
Guy Halawi, Gideon Dror, Evgeniy Gabrilovich, and Yehuda Koren. 2012.
\newblock \href {https://doi.org/10.1145/2339530.2339751} {{Large-scale
  Learning of Word Relatedness with Constraints}}.
\newblock In \emph{Proceedings of the 18th ACM SIGKDD International Conference
  on Knowledge Discovery and Data Mining}, KDD '12, pages 1406--1414, Beijing,
  China. ACM.

\bibitem[{Hill et~al.(2016)Hill, Cho, and Korhonen}]{hill2016learning}
Felix Hill, Kyunghyun Cho, and Anna Korhonen. 2016.
\newblock \href {https://doi.org/10.18653/v1/N16-1162} {{Learning Distributed
  Representations of Sentences from Unlabelled Data}}.
\newblock In \emph{Proceedings of the 2016 Conference of the North American
  Chapter of the Association for Computational Linguistics: Human Language
  Technologies}, pages 1367--1377, San Diego, California. Association for
  Computational Linguistics.

\bibitem[{Hill et~al.(2015)Hill, Reichart, and Korhonen}]{hill2015simlex}
Felix Hill, Roi Reichart, and Anna Korhonen. 2015.
\newblock \href {https://doi.org/10.1162/COLI_a_00237} {{SimLex-999: Evaluating
  Semantic Models With (Genuine) Similarity Estimation}}.
\newblock \emph{Computational Linguistics}, 41(4):665--695.

\bibitem[{Hochreiter and Schmidhuber(1997)}]{hochreiter1997lstm}
Sepp Hochreiter and J{\"u}rgen Schmidhuber. 1997.
\newblock \href {https://doi.org/10.1162/neco.1997.9.8.1735} {{Long Short-Term
  Memory}}.
\newblock \emph{Neural Computation}, 9(8):1735--1780.

\bibitem[{Hu and Liu(2004)}]{hu2004cr}
Minqing Hu and Bing Liu. 2004.
\newblock \href {https://doi.org/10.1145/1014052.1014073} {{Mining and
  Summarizing Customer Reviews}}.
\newblock In \emph{Proceedings of the Tenth ACM SIGKDD International Conference
  on Knowledge Discovery and Data Mining}, KDD '04, pages 168--177, Seattle,
  Washington. ACM.

\bibitem[{Jastrzebski et~al.(2017)Jastrzebski, Le{\'s}niak, and
  Czarnecki}]{jastrzebski2017evaluate}
Stanis{\l}aw Jastrzebski, Damian Le{\'s}niak, and Wojciech~Marian Czarnecki.
  2017.
\newblock \href {https://arxiv.org/abs/1702.02170} {{How to evaluate word
  embeddings? on importance of data efficiency and simple supervised tasks}}.
\newblock \emph{arXiv preprint arXiv:1702.02170}.

\bibitem[{Jernite et~al.(2017)Jernite, Bowman, and
  Sontag}]{jernite2017discourse}
Yacine Jernite, Samuel~R. Bowman, and David Sontag. 2017.
\newblock \href {http://arxiv.org/abs/1705.00557} {{Discourse-Based Objectives
  for Fast Unsupervised Sentence Representation Learning}}.
\newblock \emph{CoRR}, abs/1705.00557.

\bibitem[{{John D. Hunter}(2007)}]{hunter2007matplotlib}
{John D. Hunter}. 2007.
\newblock \href {https://doi.org/10.1109/MCSE.2007.55} {{Matplotlib: A 2D
  Graphics Environment}}.
\newblock \emph{{Computing in Science \& Engineering}}, 9(3):90--95.

\bibitem[{Jones et~al.(2001--)Jones, Oliphant, Peterson
  et~al.}]{jones2001scipy}
Eric Jones, Travis Oliphant, Pearu Peterson, et~al. 2001--.
\newblock \href {http://www.scipy.org/} {{SciPy: Open source scientific tools
  for Python}}.

\bibitem[{Kim et~al.(2016)Kim, Jernite, Sontag, and Rush}]{kim2016character}
Yoon Kim, Yacine Jernite, David Sontag, and Alexander Rush. 2016.
\newblock \href
  {https://www.aaai.org/ocs/index.php/AAAI/AAAI16/paper/view/12489/12017}
  {{Character-Aware Neural Language Models}}.
\newblock In \emph{Proceedings of the 30th AAAI Conference on Artificial
  Intelligence}, pages 2741--2749, Phoenix, Arizona.

\bibitem[{Kipper et~al.(2008)Kipper, Korhonen, Ryant, and
  Palmer}]{kipper2008largescale}
Karin Kipper, Anna Korhonen, Neville Ryant, and Martha Palmer. 2008.
\newblock \href {https://doi.org/10.1007/s10579-007-9048-2} {{A large-scale
  classification of English verbs}}.
\newblock \emph{Language Resources and Evaluation}, 42(1):21--40.

\bibitem[{Kiros et~al.(2015)Kiros, Zhu, Salakhutdinov, Zemel, Urtasun,
  Torralba, and Fidler}]{kiros2015skipthought}
Ryan Kiros, Yukun Zhu, Ruslan~R Salakhutdinov, Richard Zemel, Raquel Urtasun,
  Antonio Torralba, and Sanja Fidler. 2015.
\newblock \href {http://papers.nips.cc/paper/5950-skip-thought-vectors.pdf}
  {{Skip-Thought Vectors}}.
\newblock In C.~Cortes, N.~D. Lawrence, D.~D. Lee, M.~Sugiyama, and R.~Garnett,
  editors, \emph{Advances in Neural Information Processing Systems 28}, pages
  3294--3302. Curran Associates, Inc.

\bibitem[{Lample et~al.(2016)Lample, Ballesteros, Subramanian, Kawakami, and
  Dyer}]{lample2016neural}
Guillaume Lample, Miguel Ballesteros, Sandeep Subramanian, Kazuya Kawakami, and
  Chris Dyer. 2016.
\newblock \href {https://doi.org/10.18653/v1/N16-1030} {{Neural Architectures
  for Named Entity Recognition}}.
\newblock In \emph{Proceedings of the 2016 Conference of the North American
  Chapter of the Association for Computational Linguistics: Human Language
  Technologies}, pages 260--270, San Diego, California. Association for
  Computational Linguistics.

\bibitem[{Li and Roth(2002)}]{li2002trec}
Xin Li and Dan Roth. 2002.
\newblock \href {http://aclweb.org/anthology/C02-1150} {{Learning Question
  Classifiers}}.
\newblock In \emph{COLING 2002: The 19th International Conference on
  Computational Linguistics}.

\bibitem[{Ling et~al.(2015)Ling, Dyer, Black, Trancoso, Fermandez, Amir,
  Marujo, and Luis}]{ling2015finding}
Wang Ling, Chris Dyer, Alan~W Black, Isabel Trancoso, Ramon Fermandez, Silvio
  Amir, Luis Marujo, and Tiago Luis. 2015.
\newblock \href {http://aclweb.org/anthology/D15-1176} {{Finding Function in
  Form: Compositional Character Models for Open Vocabulary Word
  Representation}}.
\newblock In \emph{Proceedings of the 2015 Conference on Empirical Methods in
  Natural Language Processing}, pages 1520--1530, Lisbon, Portugal. Association
  for Computational Linguistics.

\bibitem[{Linzen et~al.(2016)Linzen, Dupoux, and
  Goldberg}]{linzen2016assessing}
Tal Linzen, Emmanuel Dupoux, and Yoav Goldberg. 2016.
\newblock \href {https://www.aclweb.org/anthology/Q16-1037} {{Assessing the
  Ability of LSTMs to Learn Syntax-Sensitive Dependencies}}.
\newblock \emph{Transactions of the Association for Computational Linguistics},
  4(1):521--535.

\bibitem[{Luong and Manning(2016)}]{luong2016achieving}
Minh-Thang Luong and Christopher~D. Manning. 2016.
\newblock \href {http://www.aclweb.org/anthology/P16-1100} {{Achieving Open
  Vocabulary Neural Machine Translation with Hybrid Word-Character Models}}.
\newblock In \emph{Proceedings of the 54th Annual Meeting of the Association
  for Computational Linguistics (Volume 1: Long Papers)}, pages 1054--1063,
  Berlin, Germany. Association for Computational Linguistics.

\bibitem[{Luong et~al.(2013)Luong, Socher, and Manning}]{luong2013better}
Thang Luong, Richard Socher, and Christopher~D. Manning. 2013.
\newblock \href {http://www.aclweb.org/anthology/W13-3512} {{Better Word
  Representations with Recursive Neural Networks for Morphology}}.
\newblock In \emph{Proceedings of the Seventeenth Conference on Computational
  Natural Language Learning}, pages 104--113, Sofia, Bulgaria. Association for
  Computational Linguistics.

\bibitem[{Marelli et~al.(2014)Marelli, Menini, Baroni, Bentivogli, bernardi,
  and Zamparelli}]{marelli2014sick}
Marco Marelli, Stefano Menini, Marco Baroni, Luisa Bentivogli, Raffaella
  bernardi, and Roberto Zamparelli. 2014.
\newblock \href
  {http://www.lrec-conf.org/proceedings/lrec2014/pdf/363_Paper.pdf} {{A SICK
  cure for the evaluation of compositional distributional semantic models}}.
\newblock In \emph{Proceedings of the Ninth International Conference on
  Language Resources and Evaluation (LREC-2014)}, Reykjavik, Iceland. European
  Language Resources Association (ELRA).

\bibitem[{McKinney(2010)}]{mckinney2010pandas}
Wes McKinney. 2010.
\newblock \href
  {https://conference.scipy.org/proceedings/scipy2010/pdfs/mckinney.pdf} {{Data
  Structures for Statistical Computing in Python}}.
\newblock In \emph{Proceedings of the 9th Python in Science Conference}, pages
  51 -- 56.

\bibitem[{Mikolov et~al.(2013)Mikolov, Sutskever, Chen, Corrado, and
  Dean}]{mikolov2013distributed}
Tomas Mikolov, Ilya Sutskever, Kai Chen, Greg~S Corrado, and Jeff Dean. 2013.
\newblock \href
  {http://papers.nips.cc/paper/5021-distributed-representations-of-words-and-phrases-and-their-compositionality.pdf}
  {{Distributed Representations of Words and Phrases and their
  Compositionality}}.
\newblock In C.~J.~C. Burges, L.~Bottou, M.~Welling, Z.~Ghahramani, and K.~Q.
  Weinberger, editors, \emph{Advances in Neural Information Processing Systems
  26}, pages 3111--3119. Curran Associates, Inc.

\bibitem[{Miyamoto and Cho(2016)}]{miyamoto2016gated}
Yasumasa Miyamoto and Kyunghyun Cho. 2016.
\newblock \href {https://aclweb.org/anthology/D16-1209} {{Gated Word-Character
  Recurrent Language Model}}.
\newblock In \emph{Proceedings of the 2016 Conference on Empirical Methods in
  Natural Language Processing}, pages 1992--1997, Austin, Texas. Association
  for Computational Linguistics.

\bibitem[{Nelson et~al.(2004)Nelson, McEvoy, and
  Schreiber}]{nelson2004behavior}
Douglas~L. Nelson, Cathy~L. McEvoy, and Thomas~A. Schreiber. 2004.
\newblock \href {https://doi.org/10.3758/BF03195588} {{The University of South
  Florida free association, rhyme, and word fragment norms}}.
\newblock \emph{Behavior Research Methods, Instruments, {\&} Computers},
  36(3):402--407.

\bibitem[{Oliphant(2015)}]{oliphant2015numpy}
Travis~E. Oliphant. 2015.
\newblock \href {http://web.mit.edu/dvp/Public/numpybook.pdf} {\emph{{Guide to
  NumPy}}}, 2nd edition.
\newblock CreateSpace Independent Publishing Platform, USA.

\bibitem[{Pang and Lee(2004)}]{pang2004subj}
Bo~Pang and Lillian Lee. 2004.
\newblock \href {http://aclweb.org/anthology/P04-1035} {{A Sentimental
  Education: Sentiment Analysis Using Subjectivity Summarization Based on
  Minimum Cuts}}.
\newblock In \emph{Proceedings of the 42nd Annual Meeting of the Association
  for Computational Linguistics (ACL-04)}.

\bibitem[{Pang and Lee(2005)}]{pang2005mr}
Bo~Pang and Lillian Lee. 2005.
\newblock \href {http://aclweb.org/anthology/P05-1015} {{Seeing Stars:
  Exploiting Class Relationships for Sentiment Categorization with Respect to
  Rating Scales}}.
\newblock In \emph{Proceedings of the 43rd Annual Meeting of the Association
  for Computational Linguistics (ACL'05)}, pages 115--124, Ann Arbor, Michigan.
  Association for Computational Linguistics.

\bibitem[{Paszke et~al.(2017)Paszke, Gross, Chintala, Chanan, Yang, DeVito,
  Lin, Desmaison, Antiga, and Lerer}]{paszke2017automatic}
Adam Paszke, Sam Gross, Soumith Chintala, Gregory Chanan, Edward Yang, Zachary
  DeVito, Zeming Lin, Alban Desmaison, Luca Antiga, and Adam Lerer. 2017.
\newblock \href {https://openreview.net/pdf?id=BJJsrmfCZ} {{Automatic
  differentiation in PyTorch}}.
\newblock In \emph{NeurIPS Autodiff Workshop}, Long Beach, California.

\bibitem[{Pennington et~al.(2014)Pennington, Socher, and
  Manning}]{pennington2014glove}
Jeffrey Pennington, Richard Socher, and Christopher Manning. 2014.
\newblock \href {https://doi.org/10.3115/v1/D14-1162} {{Glove: Global Vectors
  for Word Representation}}.
\newblock In \emph{Proceedings of the 2014 Conference on Empirical Methods in
  Natural Language Processing (EMNLP)}, pages 1532--1543, Doha, Qatar.
  Association for Computational Linguistics.

\bibitem[{Perez et~al.(2018)Perez, Strub, de~Vries, Dumoulin, and
  Courville}]{perez2018film}
Ethan Perez, Florian Strub, Harm de~Vries, Vincent Dumoulin, and Aaron
  Courville. 2018.
\newblock \href
  {https://www.aaai.org/ocs/index.php/AAAI/AAAI18/paper/view/16528} {{FiLM:
  Visual Reasoning with a General Conditioning Layer}}.
\newblock In \emph{AAAI Conference on Artificial Intelligence}, New Orleans,
  Louisiana.

\bibitem[{Perone et~al.(2018)Perone, Silveira, and
  Paula}]{perone2018evaluation}
Christian~S. Perone, Roberto Silveira, and Thomas~S. Paula. 2018.
\newblock \href {http://arxiv.org/abs/1806.06259} {Evaluation of sentence
  embeddings in downstream and linguistic probing tasks}.
\newblock \emph{CoRR}, abs/1806.06259.

\bibitem[{Peters et~al.(2018)Peters, Neumann, Iyyer, Gardner, Clark, Lee, and
  Zettlemoyer}]{peters2018elmo}
Matthew Peters, Mark Neumann, Mohit Iyyer, Matt Gardner, Christopher Clark,
  Kenton Lee, and Luke Zettlemoyer. 2018.
\newblock \href {https://doi.org/10.18653/v1/N18-1202} {Deep contextualized
  word representations}.
\newblock In \emph{Proceedings of the 2018 Conference of the North American
  Chapter of the Association for Computational Linguistics: Human Language
  Technologies, Volume 1 (Long Papers)}, pages 2227--2237, New Orleans,
  Louisiana. Association for Computational Linguistics.

\bibitem[{Radford et~al.(2018)Radford, Narasimhan, Salimans, and
  Sutskever}]{radford2018improving}
Alec Radford, Karthik Narasimhan, Tim Salimans, and Ilya Sutskever. 2018.
\newblock \href
  {https://s3-us-west-2.amazonaws.com/openai-assets/research-covers/language-unsupervised/language_understanding_paper.pdf}
  {{Improving Language Understanding by Generative Pre-Training}}.
\newblock Technical report, OpenAI.

\bibitem[{Radford et~al.(2019)Radford, Wu, Child, Luan, Amodei, and
  Sutskever}]{radford2019language}
Alec Radford, Jeffrey Wu, Rewon Child, David Luan, Dario Amodei, and Ilya
  Sutskever. 2019.
\newblock \href
  {https://d4mucfpksywv.cloudfront.net/better-language-models/language-models.pdf}
  {{Language Models are Unsupervised Multitask Learners}}.
\newblock Technical report, OpenAI.

\bibitem[{Radinsky et~al.(2011)Radinsky, Agichtein, Gabrilovich, and
  Markovitch}]{radinsky2011mturk287}
Kira Radinsky, Eugene Agichtein, Evgeniy Gabrilovich, and Shaul Markovitch.
  2011.
\newblock \href
  {http://wwwconference.org/proceedings/www2011/proceedings/p337.pdf} {{A Word
  at a Time: Computing Word Relatedness Using Temporal Semantic Analysis}}.
\newblock In \emph{Proceedings of the 20th International Conference on World
  Wide Web}, WWW '11, pages 337--346, Hyderabad, India.

\bibitem[{Rashtchian et~al.(2010)Rashtchian, Young, Hodosh, and
  Hockenmaier}]{rashtchian2010flickr8k}
Cyrus Rashtchian, Peter Young, Micah Hodosh, and Julia Hockenmaier. 2010.
\newblock \href {http://www.aclweb.org/anthology/W10-0721} {{Collecting Image
  Annotations Using Amazon's Mechanical Turk}}.
\newblock In \emph{Proceedings of the NAACL HLT 2010 Workshop on Creating
  Speech and Language Data with Amazon's Mechanical Turk}, pages 139--147, Los
  Angeles, California. Association for Computational Linguistics.

\bibitem[{van Rossum(1995)}]{rossum1995python}
Guido van Rossum. 1995.
\newblock \href {https://ir.cwi.nl/pub/5007} {{Python Tutorial}}.
\newblock Technical Report CS-R9526, Department of Computer Science, CWI,
  Amsterdam, The Netherlands.

\bibitem[{Rubenstein and Goodenough(1965)}]{rubenstein1965contextual}
Herbert Rubenstein and John~B. Goodenough. 1965.
\newblock \href {https://doi.org/10.1145/365628.365657} {{Contextual Correlates
  of Synonymy}}.
\newblock \emph{Communications of the ACM}, 8(10):627--633.

\bibitem[{Socher et~al.(2013)Socher, Perelygin, Wu, Chuang, Manning, Ng, and
  Potts}]{socher2013recursive}
Richard Socher, Alex Perelygin, Jean Wu, Jason Chuang, Christopher~D. Manning,
  Andrew Ng, and Christopher Potts. 2013.
\newblock \href {http://aclweb.org/anthology/D13-1170} {{Recursive Deep Models
  for Semantic Compositionality Over a Sentiment Treebank}}.
\newblock In \emph{Proceedings of the 2013 Conference on Empirical Methods in
  Natural Language Processing}, pages 1631--1642, Seattle, Washington.
  Association for Computational Linguistics.

\bibitem[{Subramanian et~al.(2018)Subramanian, Trischler, Bengio, and
  Pal}]{subramanian2018learning}
Sandeep Subramanian, Adam Trischler, Yoshua Bengio, and Christopher~J Pal.
  2018.
\newblock \href {https://openreview.net/pdf?id=B18WgG-CZ} {{Learning General
  Purpose Distributed Sentence Representations via Large Scale Multi-task
  Learning}}.
\newblock In \emph{International Conference on Learning Representations}.

\bibitem[{Tsvetkov et~al.(2015)Tsvetkov, Faruqui, Ling, Lample, and
  Dyer}]{tsvetkov2015evaluation}
Yulia Tsvetkov, Manaal Faruqui, Wang Ling, Guillaume Lample, and Chris Dyer.
  2015.
\newblock \href {http://aclweb.org/anthology/D15-1243} {{Evaluation of Word
  Vector Representations by Subspace Alignment}}.
\newblock In \emph{Proceedings of the 2015 Conference on Empirical Methods in
  Natural Language Processing}, pages 2049--2054, Lisbon, Portugal. Association
  for Computational Linguistics.

\bibitem[{Wang et~al.(2019)Wang, Singh, Michael, Hill, Levy, and
  Bowman}]{wang2019glue}
Alex Wang, Amanpreet Singh, Julian Michael, Felix Hill, Omer Levy, and
  Samuel~R. Bowman. 2019.
\newblock {GLUE: A Multi-Task Benchmark and Analysis Platform for Natural
  Language Understanding}.
\newblock In \emph{Proceedings of the 7th International Conference on Learning
  Representations (ICLR)}, New Orleans, Louisiana.

\bibitem[{Wang and Manning(2012)}]{wang2012baselines}
Sida Wang and Christopher Manning. 2012.
\newblock \href {http://aclweb.org/anthology/P12-2018} {{Baselines and Bigrams:
  Simple, Good Sentiment and Topic Classification}}.
\newblock In \emph{Proceedings of the 50th Annual Meeting of the Association
  for Computational Linguistics (Volume 2: Short Papers)}, pages 90--94, Jeju
  Island, Korea. Association for Computational Linguistics.

\bibitem[{Waskom et~al.(2018)Waskom, Botvinnik, O'Kane, Hobson, Ostblom,
  Lukauskas, Gemperline, Augspurger, Halchenko, Cole, Warmenhoven, de~Ruiter,
  Pye, Hoyer, Vanderplas, Villalba, Kunter, Quintero, Bachant, Martin, Meyer,
  Miles, Ram, Brunner, Yarkoni, Williams, Evans, Fitzgerald, Brian, and
  Qalieh}]{waskom2018seaborn}
Michael Waskom, Olga Botvinnik, Drew O'Kane, Paul Hobson, Joel Ostblom, Saulius
  Lukauskas, David~C Gemperline, Tom Augspurger, Yaroslav Halchenko, John~B.
  Cole, Jordi Warmenhoven, Julian de~Ruiter, Cameron Pye, Stephan Hoyer, Jake
  Vanderplas, Santi Villalba, Gero Kunter, Eric Quintero, Pete Bachant, Marcel
  Martin, Kyle Meyer, Alistair Miles, Yoav Ram, Thomas Brunner, Tal Yarkoni,
  Mike~Lee Williams, Constantine Evans, Clark Fitzgerald, Brian, and Adel
  Qalieh. 2018.
\newblock \href {https://doi.org/10.5281/zenodo.1313201} {mwaskom/seaborn:
  v0.9.0 (july 2018)}.

\bibitem[{Welch(1947)}]{welch1947generalization}
Bernard~Lewis Welch. 1947.
\newblock \href {https://doi.org/10.1093/biomet/34.1-2.28} {{The Generalization
  of ``Student's'' Problem When Several Different Population Variances are
  Involved}}.
\newblock \emph{Biometrika}, 34(1-2):28--35.

\bibitem[{Wiebe et~al.(2005)Wiebe, Wilson, and Cardie}]{wiebe2005mpqa}
Janyce Wiebe, Theresa Wilson, and Claire Cardie. 2005.
\newblock \href {https://doi.org/10.1007/s10579-005-7880-9} {{Annotating
  Expressions of Opinions and Emotions in Language}}.
\newblock \emph{Language Resources and Evaluation}, 39(2):165--210.

\bibitem[{Wieting and Kiela(2019)}]{wieting2019no}
John Wieting and Douwe Kiela. 2019.
\newblock \href {https://openreview.net/pdf?id=BkgPajAcY7} {{No Training
  Required: Exploring Random Encoders for Sentence Classification}}.
\newblock In \emph{Proceedings of the 7th International Conference on Learning
  Representations (ICLR)}, New Orleans, Louisiana.

\bibitem[{Williams et~al.(2018)Williams, Nangia, and
  Bowman}]{williams2018broad}
Adina Williams, Nikita Nangia, and Samuel Bowman. 2018.
\newblock \href {http://www.aclweb.org/anthology/N18-1101} {{A Broad-Coverage
  Challenge Corpus for Sentence Understanding through Inference}}.
\newblock In \emph{Proceedings of the 2018 Conference of the North American
  Chapter of the Association for Computational Linguistics: Human Language
  Technologies, Volume 1 (Long Papers)}, pages 1112--1122, New Orleans,
  Louisiana. Association for Computational Linguistics.

\bibitem[{Wu et~al.(2016)Wu, Schuster, Chen, Le, Norouzi, Macherey, Krikun,
  Cao, Gao, Macherey, Klingner, Shah, Johnson, Liu, Kaiser, Gouws, Kato, Kudo,
  Kazawa, Stevens, Kurian, Patil, Wang, Young, Smith, Riesa, Rudnick, Vinyals,
  Corrado, Hughes, and Dean}]{wu2016google}
Yonghui Wu, Mike Schuster, Zhifeng Chen, Quoc~V. Le, Mohammad Norouzi, Wolfgang
  Macherey, Maxim Krikun, Yuan Cao, Qin Gao, Klaus Macherey, Jeff Klingner,
  Apurva Shah, Melvin Johnson, Xiaobing Liu, Lukasz Kaiser, Stephan Gouws,
  Yoshikiyo Kato, Taku Kudo, Hideto Kazawa, Keith Stevens, George Kurian,
  Nishant Patil, Wei Wang, Cliff Young, Jason Smith, Jason Riesa, Alex Rudnick,
  Oriol Vinyals, Greg Corrado, Macduff Hughes, and Jeffrey Dean. 2016.
\newblock \href {http://arxiv.org/abs/1609.08144} {{Google's Neural Machine
  Translation System: Bridging the Gap between Human and Machine Translation}}.
\newblock \emph{arXiv preprint arXiv:1609.08144}.

\bibitem[{Yang et~al.(2017)Yang, Dhingra, Yuan, Hu, Cohen, and
  Salakhutdinov}]{yang2017words}
Zhilin Yang, Bhuwan Dhingra, Ye~Yuan, Junjie Hu, William~W. Cohen, and Ruslan
  Salakhutdinov. 2017.
\newblock \href {https://openreview.net/pdf?id=B1hdzd5lg} {{Words or
  Characters? Fine-grained Gating for Reading Comprehension}}.
\newblock In \emph{Proceedings of the 5th International Conference on Learning
  Representations (ICLR)}, Toulon, France.

\bibitem[{Young et~al.(2014)Young, Lai, Hodosh, and
  Hockenmaier}]{young2014flickr30k}
Peter Young, Alice Lai, Micah Hodosh, and Julia Hockenmaier. 2014.
\newblock \href {http://aclweb.org/anthology/Q14-1006} {{From image
  descriptions to visual denotations}}.
\newblock \emph{Transactions of the Association for Computational Linguistics},
  2:67--78.

\bibitem[{Zhang et~al.(2018)Zhang, Wu, Li, and Li}]{zhang2018learning}
Minghua Zhang, Yunfang Wu, Weikang Li, and Wei Li. 2018.
\newblock \href {http://aclweb.org/anthology/D18-1481} {{Learning Universal
  Sentence Representations with Mean-Max Attention Autoencoder}}.
\newblock In \emph{Proceedings of the 2018 Conference on Empirical Methods in
  Natural Language Processing}, pages 4514--4523, Brussels, Belgium.
  Association for Computational Linguistics.

\bibitem[{Zhao et~al.(2015)Zhao, Lu, and Poupart}]{zhao2015self-adaptive}
Han Zhao, Zhengdong Lu, and Pascal Poupart. 2015.
\newblock \href
  {https://www.aaai.org/ocs/index.php/IJCAI/IJCAI15/paper/view/10828}
  {{Self-Adaptive Hierarchical Sentence Model}}.
\newblock In \emph{Proceedings of the 24th International Joint Conference on
  Artificial Intelligence}, pages 4069--4076, Buenos Aires, Argentina. AAAI
  Press.

\bibitem[{Zhu et~al.(2018)Zhu, Li, and de~Melo}]{zhu2018exploring}
Xunjie Zhu, Tingfeng Li, and Gerard de~Melo. 2018.
\newblock \href {https://www.aclweb.org/anthology/P18-2100} {{Exploring
  Semantic Properties of Sentence Embeddings}}.
\newblock In \emph{Proceedings of the 56th Annual Meeting of the Association
  for Computational Linguistics (Volume 2: Short Papers)}, pages 632--637,
  Melbourne, Australia. Association for Computational Linguistics.

\end{thebibliography}
\bibliographystyle{acl_natbib}
\interlinepenalty=0

\appendix

\section{Hyperparameters}%
\label{app:hyperparams}

We only considered words that appear at least twice, for each dataset. Those
that appeared only once were considered UNK\@. We used the Treebank Word
Tokenizer as implemented in NLTK\footnote{\scriptsize
\url{https://www.nltk.org/}} for tokenizing the training and development
datasets.

In the same fashion as \newcite{conneau2017supervised}, we used a batch size of
64, an SGD optmizer with an initial learning rate of $0.1$, and at each epoch
divided the learning rate by $5$ if the validation accuracy decreased. We also
used gradient clipping when gradients where $> 5$.

We defined character vector representations as $50$-dimensional vectors randomly
initialized by sampling from the uniform distribution in the $(-0.05; 0.05)$
range.

The output dimension of the character-level BiLSTM was $300$ per
direction, and remained of such size after combining forward and backward
representations as depicted in eq.~\ref{eq:fw_bw_char}.

\begin{table*}[hbt!]
\scriptsize
\centering
\begin{tabular}{lll}
\toprule
\textbf{Dataset} & \textbf{Reference} & \textbf{URL}                \\
\midrule
MEN                                                & \citet{bruni2014men}              & \url{https://staff.fnwi.uva.nl/e.bruni/MEN}                         \\
MTurk287                                           & \citet{radinsky2011mturk287}      & \url{https://git.io/fhQA8} (Unofficial)                             \\
MTurk771                                           & \citet{halawi2012mturk771}        &
\url{http://www2.mta.ac.il/~gideon/mturk771.html} \\
RG                                                 & \citet{rubenstein1965contextual}  & \url{https://git.io/fhQAB} (Unofficial)                             \\
RareWords (RW)                                     & \citet{luong2013better}           & \url{https://nlp.stanford.edu/~lmthang/morphoNLM/}                  \\
SimLex999                                          & \citet{hill2015simlex}            & \url{https://fh295.github.io/simlex.html}                           \\
SimVerb3500                                        & \citet{gerz2016simverb}           & \url{http://people.ds.cam.ac.uk/dsg40/simverb.html}                 \\
WS353                                              & \citet{finkelstein2002wordsim353} & \url{http://www.cs.technion.ac.il/~gabr/resources/data/wordsim353/} \\
WS353R                                             & \citet{agirre2009study}           & \url{http://alfonseca.org/eng/research/wordsim353.html}             \\
WS353S                                             & \citet{agirre2009study}           & \url{http://alfonseca.org/eng/research/wordsim353.html}             \\
\bottomrule
\end{tabular}

\renewcommand\thetable{B.1}
\caption{Word similarity and relatedness
datasets.}\label{table:word-similarity-dataset}

\end{table*}

Word vector representations where initialized from the $300$-dimensional GloVe
vectors~\cite{pennington2014glove}, trained in 840B tokens from the Common
Crawl\footnote{\scriptsize \url{https://nlp.stanford.edu/projects/glove/}}, and
finetuned during training.  Words not present in the GloVe vocabulary where
randomly initialized by sampling from the uniform distribution in the $(-0.05;
0.05)$ range.

The input size of the word-level LSTM was $300$ for every method except
\texttt{concat} in which it was $600$, and its output was always $2048$ per
direction, resulting in a $4096$-dimensional sentence representation.

\section{Datasets}\label{appendix:datasets}

\subsection{Word Similarity}\label{appendix:word-sim-datasets}

\Cref{table:word-similarity-dataset} lists the word-similarity datasets and
their corresponding reference. As mentioned in \cref{subsec:datasets}, all the
word-similarity datasets contain pairs of words annotated with similarity or
relatedness scores, although this difference is not always explicit. Below we
provide some details for each.

\textbf{MEN} contains 3000 annotated word pairs with integer scores ranging from
0 to 50. Words correspond to image labels appearing in the
ESP-Game\footnote{\scriptsize \url{http://www.cs.cmu.edu/~biglou/resources/}}
and MIRFLICKR-1M\footnote{\scriptsize \url{http://press.liacs.nl/mirflickr/}}
image datasets.

\textbf{MTurk287} contains 287 annotated pairs with scores ranging from 1.0 to
5.0.  It was created from words appearing in both DBpedia and in news articles
from The New York Times.

\textbf{MTurk771} contains 771 annotated pairs with scores ranging from 1.0 to
5.0, with words having synonymy, holonymy or meronymy relationships sampled from
WordNet \citep{fellbaum98wordnet}.

\textbf{RG} contains 65 annotated pairs with scores ranging from 0.0 to 4.0
representing ``similarity of meaning''.

\textbf{RW} contains 2034 pairs of words annotated with similarity scores in a
scale from 0 to 10. The words included in this dataset were obtained from
Wikipedia based on their frequency, and later filtered depending on their
WordNet synsets, including synonymy, hyperonymy, hyponymy, holonymy and
meronymy. This dataset was created with the purpose of testing how
well models can represent rare and complex words. 

\textbf{SimLex999} contains 999 word pairs annotated with similarity scores
ranging from 0 to 10. In this case the authors explicitly considered similarity
and not relatedness, addressing the shortcomings of datasets that do not, such
as MEN and WS353. Words include nouns, adjectives and verbs.

\textbf{SimVerb3500} contains 3500 verb pairs annotated with similarity scores
ranging from 0 to 10. Verbs were obtained from the USF free association
database~\citep{nelson2004behavior}, and VerbNet~\citep{kipper2008largescale}.
This dataset was created to address the lack of representativity of verbs in
\mbox{SimLex999}, and the fact that, at the time of creation, the best
performing models had already surpassed inter-annotator agreement in verb
similarity evaluation resources. Like SimLex999, this dataset also explicitly
considers similarity as opposed to relatedness.

\textbf{WS353} contains 353 word pairs annotated with similarity scores from 0
to 10.

\textbf{WS353R} is a subset of WS353 containing 252 word pairs annotated with
relatedness scores. This dataset was created by asking humans to classify each
WS353 word pair into one of the following classes: synonyms, antonyms,
identical, hyperonym-hyponym, hyponym-hyperonym, holonym-meronym,
meronym-holonym, and none-of-the-above. These annotations were later used to
group the pairs into: \textit{similar} pairs (synonyms, antonyms, identical,
hyperonym-hyponym, and hyponym-hyperonym), \textit{related} pairs
(holonym-meronym, meronym-holonym, and none-of-the-above with a human similarity
score greater than 5), and \textit{unrelated} pairs (classified as
none-of-the-above with a similarity score less than or equal to 5). This
dataset is composed by the union of related and unrelated pairs.

\begin{table*}[!htbp]
\scriptsize
\centering
\begin{tabular}{lll}
\toprule
\textbf{Dataset} & \textbf{Reference} & \textbf{URL}                \\
\midrule
CR    & \citet{hu2004cr} & \url{https://www.cs.uic.edu/~liub/FBS/sentiment-analysis.html#datasets}\\
MPQA  & \citet{wiebe2005mpqa} & \url{https://mpqa.cs.pitt.edu/corpora/mpqa_corpus/}\\
MR    & \citet{pang2005mr} & \url{http://www.cs.cornell.edu/people/pabo/movie-review-data/}\\
SST2  & \citet{arora2017simple} & \url{https://github.com/PrincetonML/SIF/tree/master/data}\\
SST5  &  See caption. & \url{https://git.io/fhQAV}\\
SUBJ  & \citet{pang2004subj} & \url{http://www.cs.cornell.edu/people/pabo/movie-review-data/}\\
TREC  & \citet{li2002trec} & \url{http://cogcomp.org/Data/QA/QC/}\\
SICKE & \citet{marelli2014sick} & \url{http://clic.cimec.unitn.it/composes/sick.html}\\
SICKR & \citet{marelli2014sick} & \url{http://clic.cimec.unitn.it/composes/sick.html}\\
STS16 & \citet{agirre2016semeval} & \url{http://ixa2.si.ehu.es/stswiki/index.php/Main_Page}\\
STSB  & \citet{cer2017semeval} & \url{http://ixa2.si.ehu.es/stswiki/index.php/STSbenchmark}\\

\bottomrule
\end{tabular}

\renewcommand\thetable{B.2}
\caption{Sentence representation evaluation datasets. SST5 was obtained from a
GitHub repository with no associated peer-reviewed
work.}\label{table:sentence-eval-datasets}

\end{table*}

\textbf{WS353S} is another subset of WS353 containing 203 word pairs annotated
with similarity scores. This dataset is composed by the union of similar and
unrelated pairs, as described previously.

% The definition of similarity the authors use include notions
% of relatedness as well

\subsection{Sentence Evaluation Datasets}

\Cref{table:sentence-eval-datasets} lists the sentence-level evaluation datasets
used in this paper. The provided URLs correspond to the original sources, and
not necessarily to the URLs where SentEval\footnote{\tiny
\url{https://github.com/facebookresearch/SentEval/tree/906b34a}} got the data from\footnote{A list of
    the data used by SentEval can be found in its data setup script: \tiny
\url{https://git.io/fhQpq}}.

% All the sentence-level datasets used in this paper are automatically obtained
% and preprocessed by the SentEval software. Here we
% describe the original sources.

The version of the CR, MPQA, MR, and SUBJ datasets used in this paper were the
ones preprocessed by \citet{wang2012baselines}\footnote{\tiny
\url{https://nlp.stanford.edu/~sidaw/home/projects:nbsvm}}. Both SST2 and SST5
correspond to preprocessed versions of the \ac{SST} dataset by
\citet{socher2013recursive}\footnote{\tiny
\url{https://nlp.stanford.edu/sentiment/}}. SST2 corresponds to a subset of
\ac{SST} used by \citet{arora2017simple} containing flat representations of
sentences annotated with binary sentiment labels, and SST5 to another subset
annotated with more fine-grained sentiment labels (very negative, negative,
neutral, positive, very positive).

\end{document}